\documentclass[10pt,twocolumn,letterpaper]{article}

\usepackage{iccv}              

\definecolor{iccvblue}{rgb}{0.21,0.49,0.74}
\usepackage[pagebackref,breaklinks,colorlinks,allcolors=iccvblue]{hyperref}
\usepackage{adjustbox}
\usepackage{utfsym}

\usepackage{multirow,algorithm}
\usepackage[utf8]{inputenc} 
\usepackage[T1]{fontenc}    
\usepackage{booktabs}       
\usepackage{amsfonts}       
\usepackage{nicefrac}       
\usepackage{microtype}      
\usepackage[dvipsnames]{xcolor}
\usepackage{colortbl}
\usepackage{enumitem}
\usepackage{algpseudocode}
\usepackage{times}
\usepackage{epsfig}
\usepackage{amssymb}
\usepackage{caption}
\usepackage{subcaption}
\usepackage{tabularx}
\usepackage{rotating}
\usepackage{etoolbox}
\usepackage[normalem]{ulem}
\usepackage{lipsum}
\usepackage{stmaryrd}
\usepackage{stackengine}
\usepackage{makecell}
\usepackage{pifont}
\usepackage{cancel}
\usepackage{wrapfig}
\usepackage{bm}
\usepackage{mathtools}

\usepackage{tikz}
\definecolor{barrier}{RGB}{112,128,144}

\title{DiST-4D: Disentangled Spatiotemporal Diffusion with Metric Depth \\ for 4D Driving Scene Generation}

\author{
Jiazhe Guo$^{1,2}$\thanks{Equal contribution, $\dagger$ Project leader, $\ddagger$ Corresponding author }~ Yikang Ding$^{2* \dagger}$~  Xiwu Chen$^{3\ddagger}$~ Shuo Chen$^{1,2}$~ Bohan Li$^{2,4}$~ Yingshuang Zou$^{1,2}$ \\
Xiaoyang Lyu$^{5}$~ Feiyang Tan$^{3}$~ Xiaojuan Qi$^{5}$~ 
Zhiheng Li$^{1}$~ Hao Zhao$^{1 \ddagger}$  \\ \vspace{-11pt}\\
$^{1}$THU~~ $^{2}$MEGVII~~ $^{3}$Mach Drive~~ $^{4}$SJTU~~ $^{5}$HKU \\
\href{https://royalmelon0505.github.io/DiST-4D}{https://royalmelon0505.github.io/DiST-4D}
}

\begin{document}
\maketitle
\begin{abstract}

Current generative models struggle to synthesize dynamic 4D driving scenes that simultaneously support temporal extrapolation and spatial novel view synthesis (NVS) without per-scene optimization. A key challenge lies in finding an efficient and generalizable geometric representation that seamlessly connects temporal and spatial synthesis. To address this, we propose DiST-4D, the first disentangled spatiotemporal diffusion framework for 4D driving scene generation, which leverages metric depth as the core geometric representation. DiST-4D decomposes the problem into two diffusion processes: DiST-T, which predicts future metric depth and multi-view RGB sequences directly from past observations, and DiST-S, which enables spatial NVS by training only on existing viewpoints while enforcing cycle consistency. This cycle consistency mechanism introduces a forward-backward rendering constraint, reducing the generalization gap between observed and unseen viewpoints. Metric depth is essential for both accurate reliable forecasting and accurate spatial NVS, as it provides a view-consistent geometric representation that generalizes well to unseen perspectives. Experiments demonstrate that DiST-4D achieves state-of-the-art performance in both temporal prediction and NVS tasks, while also delivering competitive performance in planning-related evaluations. 

\end{abstract}

\vspace{-10pt}
\section{Introduction}
Generative models have recently emerged as a powerful tool for autonomous driving, enabling the creation of large-scale synthetic data to train and evaluate perception, planning, and simulation systems~\cite{gao2023magicdrive, li2024uniscene, gao2024magicdrivedit, wang2023drivewm, wang2023drivedreamer, huang2024subjectdrive, wen2023panacea}. These methods alleviate the limitations of costly real-world data collection, providing diverse and scalable training scenarios crucial for improving model robustness and generalization. However, existing generative approaches for autonomous driving remain limited in their ability to synthesize realistic, controllable open-world environments. A key yet unresolved challenge lies in generating dynamic 4D scenes (3D geometry + temporal evolution) that can support data generation at any novel time and location.

\begin{table}[!t]
\vspace{-0pt}
\begin{center}
\scriptsize
\vspace{-0pt}
\renewcommand\tabcolsep{7.5pt}
\centering
\resizebox{1.0\linewidth}{!}{
\begin{tabular}{l|ccc}
\toprule Method & Temporal Gen. & Spatial NVS & Feed-forward  \\
\midrule
Vista~\cite{gao2024vista}  &  \textcolor{ForestGreen}{\usym{2713}}  &  \textcolor{red}{\usym{2717}}   & \textcolor{ForestGreen}{\usym{2713}}  \\
UniScene~\cite{li2024uniscene}   & \textcolor{ForestGreen}{\usym{2713}}   & \textcolor{red}{\usym{2717}}  & \textcolor{ForestGreen}{\usym{2713}}  \\
Drive-WM~\cite{wang2023drivewm}  &  \textcolor{ForestGreen}{\usym{2713}}  &  \textcolor{red}{\usym{2717}}  &  \textcolor{ForestGreen}{\usym{2713}}  \\
DriveDreamer~\cite{wang2023drivedreamer}  & \textcolor{ForestGreen}{\usym{2713}}  &  \textcolor{red}{\usym{2717}}   &  \textcolor{ForestGreen}{\usym{2713}}  \\
MagicDrive~\cite{gao2023magicdrive}  & \textcolor{ForestGreen}{\usym{2713}}  & \textcolor{red}{\usym{2717}}   & \textcolor{ForestGreen}{\usym{2713}}   \\
MagicDriveDiT~\cite{gao2024magicdrivedit}  &  \textcolor{ForestGreen}{\usym{2713}} & \textcolor{red}{\usym{2717}}   &  \textcolor{ForestGreen}{\usym{2713}}  \\
\midrule
3DGS~\cite{gaussian_splatting}  & \textcolor{red}{\usym{2717}}  & \textcolor{ForestGreen}{\usym{2713}} & \textcolor{red}{\usym{2717}}  \\
PVG~\cite{chen2023PVG}  & \textcolor{red}{\usym{2717}}  & \textcolor{ForestGreen}{\usym{2713}} & \textcolor{red}{\usym{2717}}  \\
UniSim~\cite{unisim}  & \textcolor{red}{\usym{2717}}  & \textcolor{ForestGreen}{\usym{2713}} & \textcolor{red}{\usym{2717}}  \\
EmerNeRF~\cite{yang2023emernerf}   & \textcolor{red}{\usym{2717}}  & \textcolor{ForestGreen}{\usym{2713}} &  \textcolor{red}{\usym{2717}}  \\
StreetGaussian~\cite{yan2024street}  & \textcolor{red}{\usym{2717}}  & \textcolor{ForestGreen}{\usym{2713}} & \textcolor{red}{\usym{2717}}  \\
DriveDreamer4D~\cite{zhao2024drivedreamer4d}   & \textcolor{red}{\usym{2717}}  &  \textcolor{ForestGreen}{\usym{2713}} &  \textcolor{red}{\usym{2717}}  \\
\midrule
Stag-1~\cite{wang2024stag}    & \textcolor{red}{\usym{2717}}  & \textcolor{ForestGreen}{\usym{2713}} & \textcolor{ForestGreen}{\usym{2713}} \\
FreeVS~\cite{wang2024freevs}      & \textcolor{red}{\usym{2717}}  & \textcolor{ForestGreen}{\usym{2713}} & \textcolor{ForestGreen}{\usym{2713}} \\
STORM~\cite{yang2024storm}    & \textcolor{red}{\usym{2717}}  & \textcolor{ForestGreen}{\usym{2713}} & \textcolor{ForestGreen}{\usym{2713}} \\
DriveForward~\cite{tian2024drivingforward}   & \textcolor{red}{\usym{2717}}  & \textcolor{ForestGreen}{\usym{2713}} & \textcolor{ForestGreen}{\usym{2713}} \\
StreetCrafter~\cite{yan2024streetcrafter}   & \textcolor{red}{\usym{2717}}  & \textcolor{ForestGreen}{\usym{2713}} & \textcolor{ForestGreen}{\usym{2713}} \\

\midrule
DreamDrive~\cite{mao2024dreamdrive}       & \textcolor{ForestGreen}{\usym{2713}} & \textcolor{ForestGreen}{\usym{2713}} & \textcolor{red}{\usym{2717}}  \\
MagicDrive3D~\cite{gao2024magicdrive3d}    & \textcolor{ForestGreen}{\usym{2713}} & \textcolor{ForestGreen}{\usym{2713}} & \textcolor{red}{\usym{2717}}  \\

\midrule
\rowcolor{gray!10}DiST-4D (Ours) & \textcolor{ForestGreen}{\usym{2713}}&  \textcolor{ForestGreen}{\usym{2713}} & \textcolor{ForestGreen}{\usym{2713}} \\ 

\bottomrule
\end{tabular}
 }
\caption{\textbf{Comparision of different methods.} DiST-4D is the first framework to achieve feed-forward dynamic 4D driving scene generation with both temporal extrapolation and spatial novel view synthesis. This is made possible by its disentangled spatiotemporal diffusion design with metric depth.
}
\vspace{-20pt}
\label{teaser_img}
\end{center}
\end{table}

To better understand the limitations of existing approaches, we briefly review current methods for autonomous driving scene generation, as shown in Tab.\ref{teaser_img}. (1) Some methods leverage diffusion models conditioned on BEV layouts or trajectories to predict future scenarios, generating 2D videos~\cite{li2024uniscene, gao2024vista, gao2023magicdrive, gao2024magicdrivedit, mao2024dreamdrive, wang2023drivedreamer, wen2023panacea, lu2023wovogen}, 3D occupancy maps~\cite{wang2024occsora, wei2024occllama, zheng2023occworld, li2024uniscene}, or LiDAR data~\cite{zyrianov2024lidardm, li2024uniscene}. However, these approaches remain tightly coupled with predefined trajectories, preventing them from achieving novel view synthesis (NVS) capabilities.
(2) Meanwhile, implicit reconstruction paradigms ~\cite{yan2024street, yang2023emernerf, chen2023PVG, chen2024omnire} excel in NVS, but their reliance on per-scene optimization results in high computational costs. (3) Recent feed-forward solutions~\cite{yang2024storm, wang2024stag, wang2024freevs, tian2024drivingforward} employ large models to synthesize novel views from captured sensor data but lack temporal extrapolation, limiting their ability to predict future scene evolution. (4) Some emerging frameworks attempt to bridge temporal prediction and spatial NVS by reconstructing generated video sequences~\cite{gao2024magicdrive3d, mao2024dreamdrive}. However, these hybrid approaches still inherit fundamental limitations from per-scene optimization, restricting their efficiency and scalability. This highlights a critical open challenge: how to achieve simultaneous anytime future forecasting and anywhere NVS without per-scene optimization.

To tackle these challenges, we posit that the key to simultaneously achieving temporal evolution and spatial NVS lies in establishing appropriate 4D representations.
Such a representation must satisfy two critical requirements:
(1) It should encode geometric information sufficient for NVS at each time step, and (2) It must be learnable by generative models to support temporal prediction.
Failing to meet the first requirement would necessitate per-scene reconstruction, making feed-forward generation infeasible. Meanwhile, failing to meet the second criterion would prevent future state generation.
Given these considerations, along with the inherent complexity of driving scenes that challenge single-stage 4D representation learning, we propose per-frame metric depth as the representation to enable disentangled spatiotemporal 4D generation.

This choice offers three key advantages: (1) Metric depth simultaneously satisfies the two aforementioned criteria, \textit{i.e.}, providing accurate geometric information while being easily learnable by generative models~\cite{ke2023repurposing, fu2024geowizard}, (2) It exhibits strong generalization for novel view synthesis, as evidenced in~\cite{yu2024viewcrafter,wang2024freevs}, and (3) Dense metric depth itself holds significant practical value in autonomous driving applications, such as data annotation and training of perception models.

Building on these insights, we propose DiST-4D, a novel framework that leverages temporally and spatially disentangled modeling to achieve feed-forward future scene generation and spatial NVS for the first time.
To enable this paradigm, we first develop a data curation pipeline for generating large-scale multi-view metric depth data. This pipeline integrates visual reconstruction~\cite{ding2022transmvsnet}, LiDAR point cloud fusion, and semantic segmentation~\cite{jain2023oneformer} to create dense 3D point cloud prompts, followed by a depth completion network~\cite{liu2024depthlab} to produce high-precision metric depth maps.

We then design DiST-4D as a dual-branch diffusion architecture, consisting of DiST-T (temporal prediction) and DiST-S (spatial synthesis).
Specifically, DiST-T generates future multi-view RGB and metric depth sequences by conditioning on historical observations and control signals (\eg, BEV maps and trajectories).
DiST-S synthesizes dense RGB and depth outputs from novel viewpoints, using a diffusion model conditioned on sparse projections from surrounding point clouds.
A key challenge in training DiST-S on real-world driving data is the limited trajectory diversity, which hinders generalization to novel out-of-distribution viewpoints. To address this, we propose a novel self-supervised cycle consistency (SCC) strategy, inspired by~\cite{zhou2016learning,zhu2017unpaired,wang2019learning}, to enhance the robustness of spatial NVS.
Specifically, we first randomly generate novel trajectories, then use trained DiST-S to generate the corresponding RGB-D sequences, which will be reprojected back to the original viewpoints to form self-supervised training pairs. This SCC strategy enables self-supervised fine-tuning, improving DiST-S's adaptability to various trajectories.

Our contributions are summarized as follows:\looseness=-1

\begin{itemize}

    \item We propose DiST-4D, the first model to simultaneously achieve temporal future prediction and spatial novel view synthesis (NVS) without requiring per-scene optimization.
    
    \item We introduce metric depth as the geometric bridge between temporal forecasting and spatial NVS, enabling efficient and generalizable 4D scene generation.

    \item We develop a multi-view metric depth curation pipeline, to provide high-quality supervision for 4D scene generation.

    \item We propose a self-supervised cycle consistency (SCC) strategy to enhance the generalization ability of DiST-S, improving spatial NVS on out-of-distribution trajectories.
    
    \item Extensive experiments demonstrate that DiST-4D outperforms existing approaches in both temporal prediction and spatial novel view synthesis, while maintaining competitive performance in planning-related evaluations. 

\end{itemize}

\section{Related Work}

\subsection{Temporal Driving Video Generation}

Recent progress in temporal driving video generation has advanced realistic, controllable autonomous driving simulations~\cite{gao2023magicdrive,gao2025vista,li2024hierarchical,gao2024magicdrivedit,li2023bridging,wang2024driving,mao2024dreamdrive,wang2024stag}.
MagicDrive\cite{gao2023magicdrive} achieves high-fidelity street-view synthesis through tailored encoding and cross-view attention mechanisms for precise 3D geometry control.
UniScene\cite{li2024uniscene} unifies multi-modal data generation (semantic occupancy, video, LiDAR) via progressive processes, improving downstream task performance with intermediate representations. Extensions like MagicDriveDiT \cite{gao2024magicdrivedit} and MagicDrive3D~\cite{gao2024magicdrive3d} address scalability challenges using DiT architectures and deformable Gaussian splatting with monocular depth initialization, respectively, enabling high-resolution 3D scene reconstruction.
DreamDrive\cite{mao2024dreamdrive} synthesizes 4D spatial-temporal scenes via hybrid Gaussian representations, balancing visual quality and generalizability.
Despite these advancements, existing methods struggle to reconcile temporal video generation with spatial novel view synthesis, limiting scalability for autonomous driving applications. To address this, we propose a disentangled spatiotemporal framework that leverages metric depth as a geometric bridge to harmonize spatial and temporal coherence.\looseness=-1

\begin{figure*}[!t]
    \centering
    \vspace{-25pt}
    \includegraphics[width=1.0\linewidth]{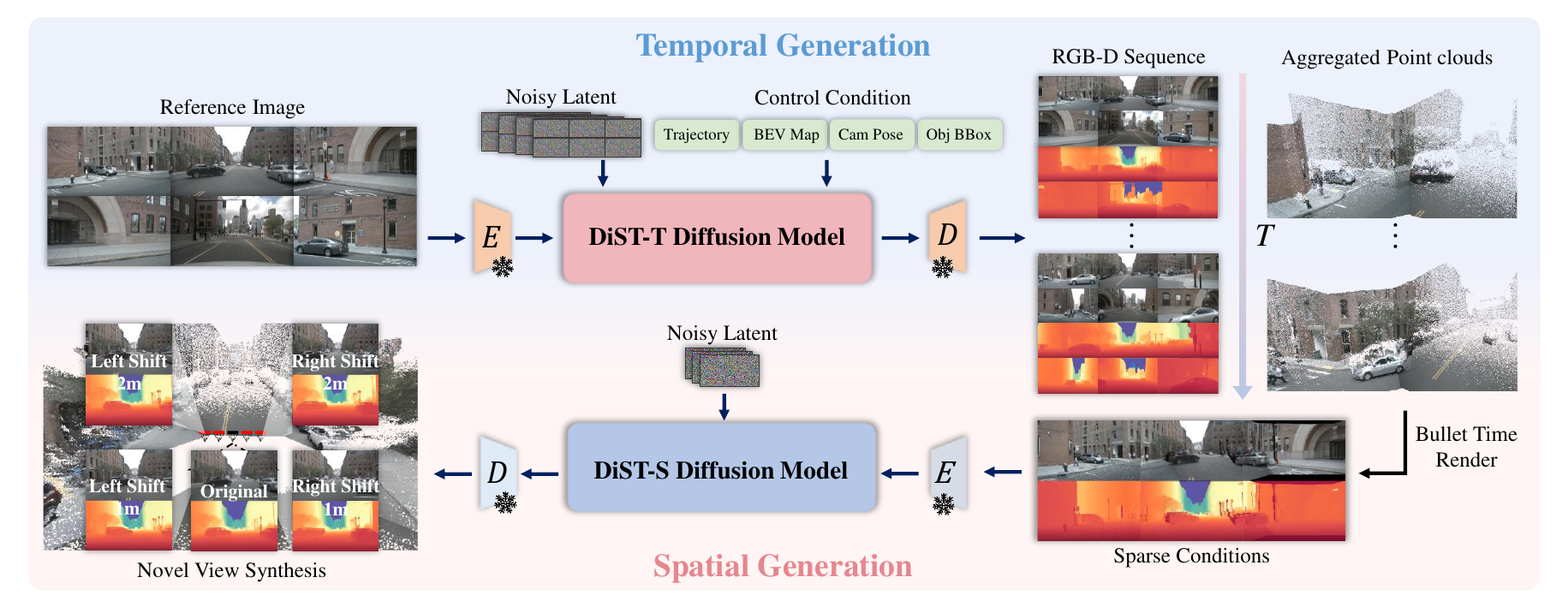}
    \vspace{-23pt}
    \caption{\textbf{Overall framework of the proposed DiST-4D.} DiST-4D is a disentangled spatiotemporal diffusion framework for 4D driving scene generation, leveraging metric depth as the core geometric representation to enable both temporal extrapolation and spatial novel view synthesis (NVS). \textbf{(Top: Temporal Generation)} DiST-T employs a diffusion model to predict future multi-camera RGB-D sequences from historical multi-camera images and control signals.
    The generated RGB-D sequences are then aggregated into point clouds, allowing for bullet time rendering. \textbf{(Bottom: Spatial Generation)} To enable spatial NVS, DiST-S leverages the predicted RGB-D sequences to generate novel viewpoints by first projecting them into sparse conditions and then refining them into dense RGB-D outputs.
    }
    \label{fig_overall}
    \vspace{-10pt}
\end{figure*}

\subsection{Spatial Novel View Synthesis}
Recent advances in spatial novel view synthesis (NVS)~\cite{tancik2022block,zhang2024gaussian,gaussian_splatting,chen2023PVG,unisim,yang2023emernerf,yan2024street,zhao2024drivedreamer4d,yang2024storm,wang2024freevs,tian2024drivingforward,yan2024streetcrafter,wang2024stag} have enhanced the generation of realistic, flexible views for dynamic driving scenes, a critical capability for autonomous driving simulations.

\noindent\textbf{Per-scene Optimizations.}
Per-scene optimization methods~\cite{gaussian_splatting,chen2023PVG,unisim,yang2023emernerf,yan2024street,zhao2024drivedreamer4d} refine scene representations to capture dynamic elements and structural details. For instance, 3D Gaussian Splatting \cite{gaussian_splatting} enables real-time radiance field rendering using Gaussian representations, avoiding computations in empty space. Periodic Vibration Gaussian (PVG)\cite{chen2023PVG} extends this framework to dynamic urban scenes by integrating temporal dynamics for unsupervised reconstruction. UniSim \cite{unisim} models static and dynamic actors via neural feature grids for multi-sensor safety-critical simulations, while EmerNeRF\cite{yang2023emernerf} decomposes scenes into static/dynamic fields using self-supervised flow. Street Gaussian \cite{yan2024street} combines explicit point clouds and 4D spherical harmonics for efficient rendering.

\noindent\textbf{Feed-forward Approaches.}
Feed-forward approaches ~\cite{yang2024storm,wang2024freevs,tian2024drivingforward,yan2024streetcrafter,wang2024stag} emphasize rapid inference and scalability for real-time applications. STORM \cite{yang2024storm} leverages transformer-based architectures to infer dynamic 3D scenes from sparse inputs, achieving high-fidelity rendering. FreeVS\cite{wang2024freevs} generates 3D-consistent novel views using pseudo-image priors, bypassing per-scene optimization.
Stag-1 \cite{wang2024stag} decouples spatiotemporal relationships to enable multi-view 4D consistency.
In this paper, we propose an innovative feed-forward 4D driving scene paradigm to efficiently achieve unified 4D dynamic driving scene generation without per-scene optimization.\looseness=-1

\section{Methodology}
\subsection{Overview}

As illustrated in Fig. \ref{fig_overall}, DiST-4D comprises two key components: DiST-T and DiST-S, designed to decouple temporal prediction from spatial novel view synthesis.
Given a historical frame of observations (\textit{i.e.}, multi camera images) and control signals (\textit{i.e.}, trajectories, object BBoxes, and BEV maps), DiST-T leverages a diffusion transformer network to generate future multi camera RGB videos and depth sequences.
For spatial NVS, the generated RGB and depth maps are projected onto novel views, forming sparse RGB-D conditions that guide DiST-S in producing dense RGB-D outputs.
Since accurate and dense metric depth is crucial, we first introduce the pipeline for obtaining metric depth for model training (Sec. \ref{subsec:depth}).
We then detail the temporal generation model, DiST-T (Sec. \ref{subsec:DiST-T}), followed by the spatial novel view synthesis model, DiST-S (Sec. \ref{subsec:DiST-S}).

\begin{figure}[!tbph]
    \centering
        \vspace{-6pt}
    \includegraphics[width=1\linewidth]{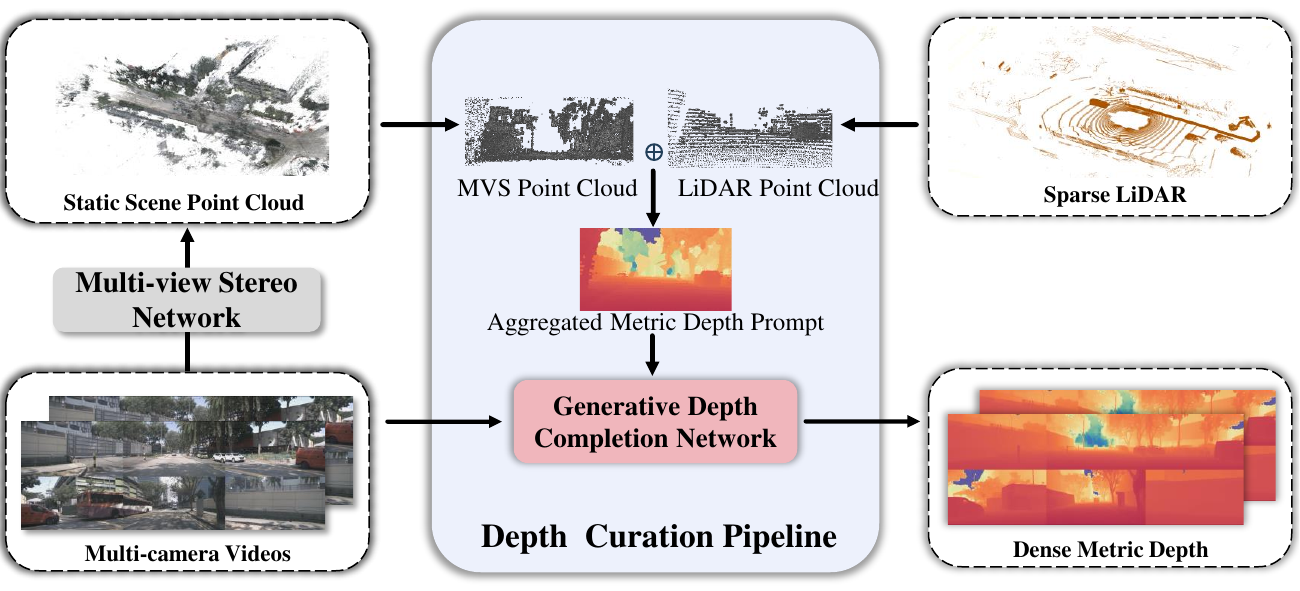}
    \vspace{-16pt}
    \caption{\textbf{Illustration of the metric depth curation pipeline.} First, a multi-view stereo network processes multi-camera videos to produce a static scene point cloud. Simultaneously, sparse LiDAR point clouds are collected and fused with the MVS output to obtain an aggregated metric depth prompt. This intermediate depth representation serves as input to a generative depth completion network, which refines and densifies the depth estimates, producing high-fidelity dense metric depth maps.}
    \label{fig_DepthGen_fw}
    \vspace{-8pt}
\end{figure}

\subsection{Metric Depth Curation}
\label{subsec:depth}
Acquiring metric depth in autonomous driving remains a challenging task. While learning-based models~\cite{wei2023surrounddepth, zou2024m2depth} have made remarkable progress in inferring metric depth from RGB images, they still suffer from inherent limitations in precision and detail fidelity. This makes them inadequate for training generative models that demand high-quality supervision.
To address this problem, we propose a pseudo-GT curation pipeline to generate high-quality metric depth.

Inspired by the recent generative depth completion method~\cite{liu2024depthlab}, we propose to get dense depth maps conditioned on RGB images and incomplete depth data, which can be obtained from LiDAR point clouds.
While LiDAR point clouds offer precise sparse depth measurements, they fall short in reconstructing complete scene geometry,
especially in capturing high buildings with upper parts being far from the ground, and distant objects located far away from the ego vehicle.
To this end, we additionally combine visual reconstruction results with LiDAR point clouds.

As shown in Fig.\ref{fig_DepthGen_fw}, we first aggregate the sparse LiDAR points extracted from adjacent frames, where dynamic points are removed using 3D bounding boxes.
Next, we obtain static scene points using an MVS network\cite{ding2022transmvsnet}, filtering out dynamic objects and the sky region via 2D semantic segmentation~\cite{jain2023oneformer}.
We then derive the initial metric depth by merging the MVS depth and LiDAR depth, followed by applying a hidden point removal algorithm~\cite{katz2007direct} to eliminate occluded points. 
Finally, the aggregated metric depth prompt, along with the corresponding RGB images, are fed into a generative depth completion network~\cite{liu2024depthlab} to produce the final dense depth maps.
More details about metric depth curation are provided in supplemental material.

\subsection{DiST-T: Temporal RGB-D Generation}
\label{subsec:DiST-T}

Given a frame of historical RGB observations as reference images $\mathbf{I}_{\text{ref}}=\{\mathbf{I}_{\text{ref},c}\}_{c=1}^{C}$, where \(C\) denotes the number of camera views, DiST-T aims to predict the future scenes corresponding to $\mathbf{I}_{\text{ref}}$ and control signals.
Similar to previous works~\cite{gao2023magicdrive, gao2024magicdrivedit}, the control signals are defined as descriptions $\{\mathbf{S}_t\}_{t=1}^{T}$ of future frames.
For each $\mathbf{S}_{t}=\{\mathbf{M}_{t}, \mathbf{B}_{t}, \mathbf{A}_{t}, \mathbf{P}_{t}\}$, \(\mathbf{M}_{t}\in\{0, 1\}^{w\times h\times s}\) represents a \(w\times h\) road region with \(s\) semantic classes. $\mathbf{B}_{t}=\{(c_i, \mathbf{b}_i)\}_{i=1}^{N}$ are 3D bounding boxes where each object is described by
the corner points of the box \(\mathbf{b}_i\in\mathbb{R}^{8\times 3}\) and class \(c_i\in\mathcal{C}\).
\(\mathbf{A}_{t}\) represents ego vehicle trajectory, describing the relative transformation from the $t$-th frame to the first frame of the sequence, and \(\mathbf{P}_{t}\in \mathbb{R}^{c\times4 \times 4}\) indicates the pose of $c$ cameras.
As we treat metric depth as the key to 4D scene generation, DiST-T needs to generate future RGB videos and depth sequences.
Let $\mathbf{Z}^{I}_{t}$ and $\mathbf{Z}_{t}^{D}$ denote the latent feature of RGB image and depth at time $t$, then DiST-T can be formulated as:
\begin{equation}
     \{\mathbf{Z}_{t}^{I},\mathbf{Z}_{t}^{D}\} = \mathcal{G}_{\theta}(\{\mathbf{S}_{t}, \mathbf{I}_{\text{ref}}\}),
\end{equation}
where $\mathcal{G}_{\theta}$ indicates DiST-T network with parameters $\theta$.

\begin{figure}[!t]
    \centering
    \includegraphics[width=1\linewidth]{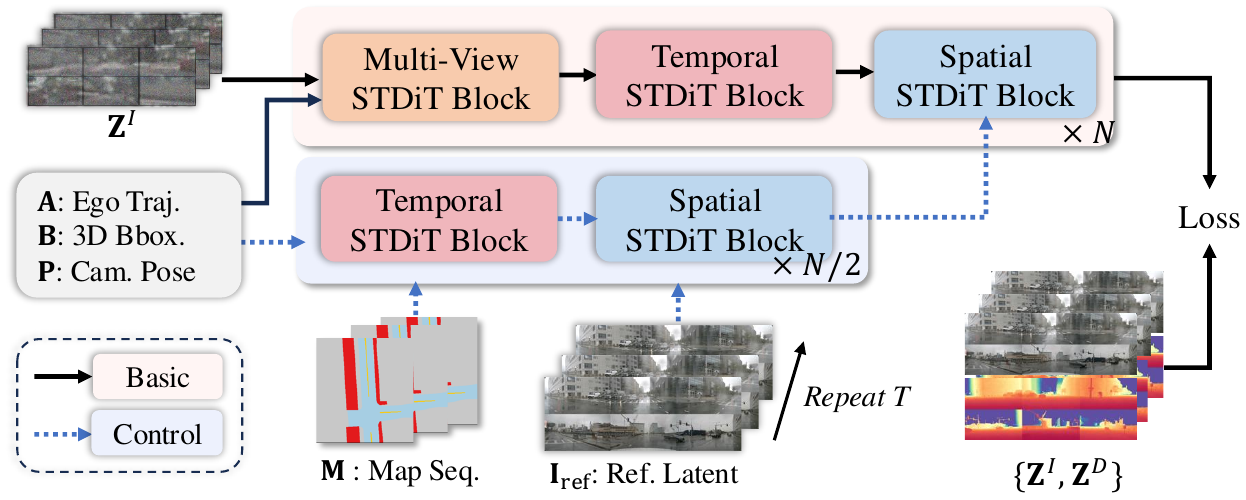}
    \vspace{-16pt}
    \caption{\textbf{Illustration of the diffusion transformer of DiST-T.} STDiT captures multi-camera and multi-frame dependencies for future scene generation. It processes latent features with temporal and spatial STDiT blocks, integrating ego trajectory ($\mathbf{A}$), 3D object bounding boxes ($\mathbf{B}$), camera poses ($\mathbf{P}$) and map sequences ($\mathbf{M}$) as control signals. 
    }
    \vspace{-6pt}
    \label{dit_framework}
\end{figure}

\paragraph{Network Architecture.}
Simultaneously generating RGB-D videos for multi-camera poses significant challenges to GPU memory.
To encode the RGB-D videos into highly compressed latent features, we utilize a pre-trained 3D VAE~\cite{yang2024cogvideox} to process RGB and depth videos separately, reducing their size in both temporal and spatial dimensions.
For the latent diffusion model, STDiT~\cite{zheng2024opensora} is employed as the backbone of DiST-T. The down-sample and embedding strategies for $\mathbf{B},\mathbf{A}$ and $\mathbf{M}$ are adopted from \cite{gao2024magicdrivedit}.
As illustrated in Fig.~\ref{dit_framework}, the control branch (highlighted in blue) is similar to ControlNet~\cite{zhang2023adding}. Map $\mathbf{M}$ and reference image $\mathbf{I}_{\text{ref}}$ are injected into the control blocks after encoding, while the $\mathbf{A}$, $\mathbf{B}$, and $\mathbf{P}$ are provided as input to both the basic block and the control block.
Within the STDiT backbone, the multi-view block captures interactions among multiple cameras, whereas the temporal and spatial blocks focus on temporal and spatial interactions within a single camera, respectively.

\paragraph{Training Strategy.}

Training a multi-modal generative model from scratch results in slow convergence. To speed up training, we first train a low-resolution ($224 \times 400$) RGB video generation model, then extend it to low-resolution RGB-D generation, and finally refine it for high-resolution ($424 \times 800$) RGB-D video generation with a mixed-frame-length \([9,17]\) strategy.
The model is trained through simulation-free
rectified flow and v-prediction loss~\cite{esser2024scaling}, more details about loss function can be found in supplementary material.

\subsection{DiST-S: Spatial RGB-D NVS}
\label{subsec:DiST-S}
\paragraph{Task Formulation.}

The DiST-S aims at synthesizing spatial novel view RGB images ${\mathbf{I}_{\text{tgt}}}$ and the corresponding depth maps ${\mathbf{D}_{\text{tgt}}}$ in a feed-forward manner.
Unlike FreeVS~\cite{wang2024freevs}, which uses sparse LiDAR data, and ViewCrafter~\cite{yu2024viewcrafter}, which relies on relative depth projections, DiST-S utilizes metric depth projections as inputs, including projected RGB and depth.
As depicted in Fig. \ref{fig_overall}, 
given an arbitrary future moment $t$, we convert the generated mutli-view RGB images $\mathbf{I}_{\text{gen}}$ and dense depth maps $\mathbf{D}_{\text{gen}}$ into a per-frame point cloud using the known camera intrinsics and extrinsics.
After that, we project point cloud at $t$ into novel viewpoint to get condition results $\mathbf{I}_{\text{cond}}$ and $\mathbf{D}_{\text{cond}}$.
Let $\mathbf{Z}^{I}_{\text{cond}}$ and $\mathbf{Z}^{D}_{\text{cond}}$ denote the latent feature of $\mathbf{I}_{\text{cond}} $ and $\mathbf{D}_{\text{cond}}$ from the frozen VAE decoder, then DiST-S can be formulated as:
\begin{equation}
     \{\mathbf{Z}^{I}_{\text{tgt}},\mathbf{Z}^{D}_{\text{tgt}}\} = \mathcal{F}_{\theta}(\{\mathbf{Z}^{I}_{\text{cond}}, \mathbf{Z}^{D}_{\text{cond}}\}),
\end{equation}
where $\mathcal{F}_{\theta}$ indicates DiST-S network with parameters $\theta$.

\begin{figure}[!t]
    \centering
    \includegraphics[width=0.99\linewidth]{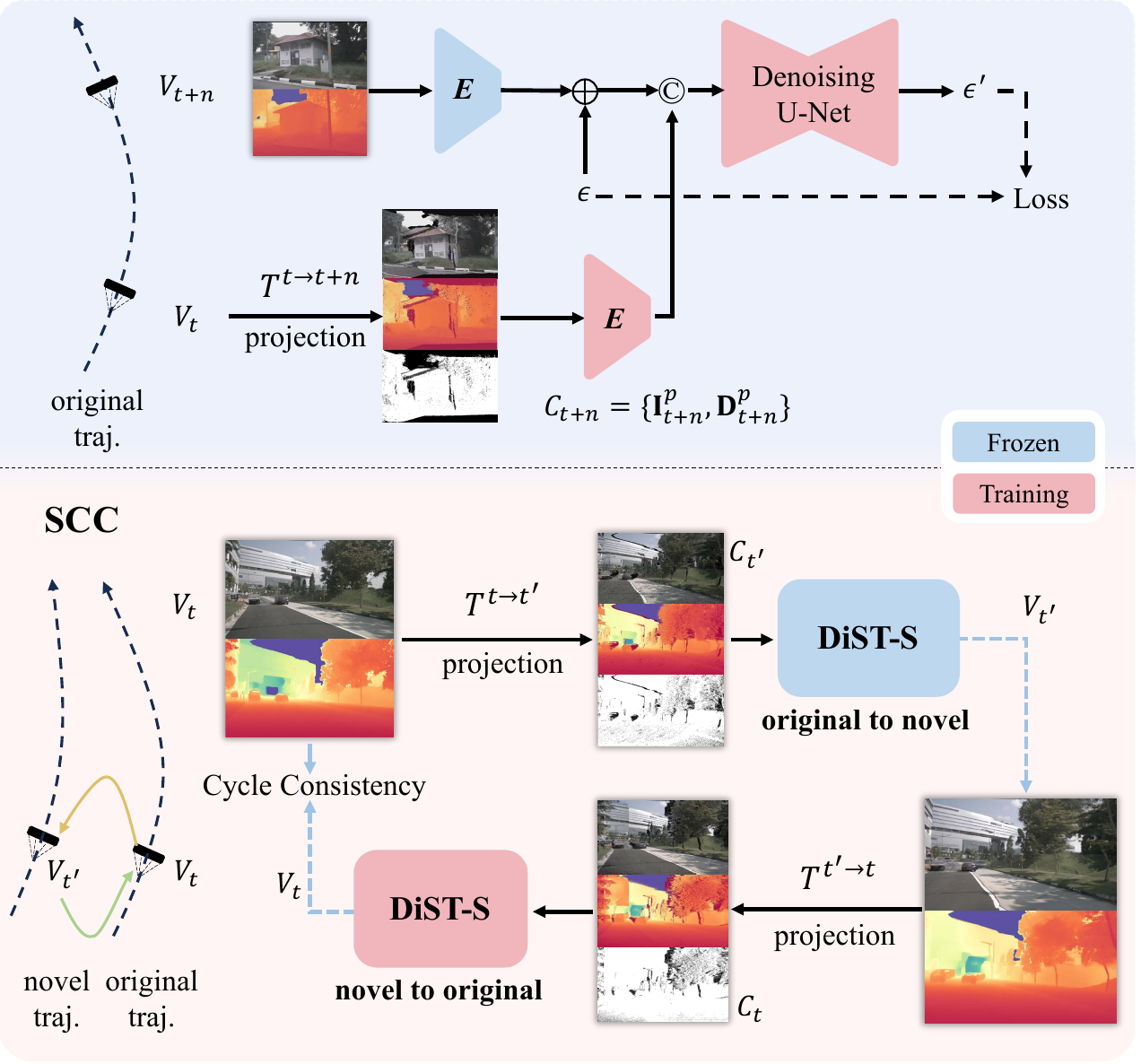}
    \vspace{-5pt}
    \caption{(Top) In the first stage, DiST-S is trained on the original trajectory using $n$ frame projection
    (Bottom) In the second stage, self-supervised cycle consistency (SCC) is introduced, where novel trajectories are generated, and DiST-S learns to project between original and novel viewpoints.
    }
    \label{unet_framework}
\end{figure}

\paragraph{Network Architecture.}
As shown in Fig.~\ref{unet_framework}, DiST-S consists of a denoising UNet~\cite{blattmann2023svd}, a frozen VAE encoder, and a trainable condition encoder.
The frozen encoder compresses the dense target RGB images and depth maps into latent features, while the trainable encoder is used to extract latent features of the sparse condition maps.

Since the VAE encoder is designed for 3-channel RGB input, we replicate the single-channel depth map into three channels to simulate an RGB image. During training, the RGB latent $\mathbf{Z}^{I}_{\text{tgt}}$ and depth latent $\mathbf{Z}^{D}_{\text{tgt}}$ are concatenated along the channel dimension. For the condition inputs, we combine the single-channel valid mask with the channel-wise replicated two-channel depth input to create the final depth condition $\mathbf{D}_\text{cond}$. Both $\mathbf{D}_\text{cond}$ and the image condition $\mathbf{I}_\text{cond}$ are then encoded into $\mathbf{Z}^{D}_\text{cond}$ and $\mathbf{Z}^{I}_\text{cond}$ through a condition encoder, which is trained from scratch. Finally, $\mathbf{Z}^{I}_\text{cond}$ and $\mathbf{Z}^{D}_\text{cond}$ are concatenated with $\mathbf{Z}^{I}_\text{tgt}$ and $\mathbf{Z}^{D}_\text{tgt}$ along the channel dimension to form the denoising inputs.

As our denoising UNet is based on the video diffusion model~\cite{blattmann2023svd}, we need to modify its original input and output channels to accommodate multi-modal generation. Following the approach in~\cite{ke2023repurposing}, we increase the input channels from 8 to 16 by replicating the input layer’s weight tensor and scaling its values by half. Similarly, we modify the output channels from 4 to 8 using the same weight adjustment.


\vspace{-12pt}
\paragraph{Training Strategy and Cycle Consistency.}
The training phase of DiST-S includes two stages. In the first stage, we use the training data from the original trajectories to train the denoising UNet and condition encoder.
To obtain condition inputs, we use projection matrix $T^{t \rightarrow t+n}$ to project the surround point cloud at time $t$ to time $t+n$ to serve as conditions $C_{t+n}=\{\mathbf{I}_{t+n}^p,\mathbf{D}_{t+n}^p\}$, with the $n$ is set to $\pm2$.  While $n=\pm4$ and voxel-down-sampling for surround point cloud is utilized as data augmentation.
By incorporating data from different cameras, DiST-S is capable of learning camera motions under various pose changes, where the data from the forward-facing camera enables the model to learn how to move forward and backward, while the data from the side-facing camera teaches the model to move laterally.

However, relying solely on real trajectory data for NVS remains insufficient to address the full range of diverse and unseen trajectories. In the second training stage, inspired by~\cite{zhou2016learning,zhu2017unpaired,wang2019learning}, we introduce a \textbf{S}elf-supervised \textbf{C}ycle \textbf{C}onsistency (SCC) strategy to enhance the model's robustness in synthesizing novel views under extreme viewpoint variations.
Specifically, we first randomly generate novel trajectories $\mathbf{A}_\text{novel}$ with the corresponding input conditions based on the original trajectories $\mathbf{A}_\text{ori}$ and surrounding point clouds. For viewpoint $V_t$ at $t$ frame in $\mathbf{A}_\text{ori}$, we randomly shift $\tau \in [-3m,+3m]$ laterally to get the novel viewpoint $V_t^{'}$ in $\mathbf{A}_\text{novel}$.
Then, we use the trained DiST-S from the first training stage to synthesize dense RGB-D outputs in $\mathbf{A}_\text{novel}$, which will be reprojected to $\mathbf{A}_\text{ori}$ to produce condition inputs.
These condition maps together with the original dense RGB-D maps in $\mathbf{A}_\text{ori}$ form the additional training pairs in the second training stage.
The SCC strategy allows the DiST-S to encounter a diverse range of new trajectories during training, significantly enhancing its generalization and robustness to novel perspectives.
Following~\cite{blattmann2023svd}, the training objective in both training stages is v-prediction based MSE loss, more details about training loss can be found in the supplementary material.

\section{Experiments}
Our experiments are conducted on nuScenes dataset~\cite{caesar2020nuscenes}. We interpolate the 2Hz keyframe annotation in a higher frame rate of 12Hz, similar to previous studies~\cite{wang2022asap,gao2023magicdrive}. We adhere to the official splits, using 700 videos for training and 150 for validation.
Further details about model setup are provided in the supplementary material.

\subsection{Temporal Generation}

\begin{table}[!t]
\vspace{-0pt}
\begin{center}
\scriptsize
\vspace{-0pt}
\renewcommand\tabcolsep{5.5pt}
\centering
\resizebox{1.0\linewidth}{!}{
\begin{tabular}{l|ccc|cc}
\toprule Method & Multi-view & Video & Depth & FID $\downarrow$ & FVD $\downarrow$ \\ \midrule
BEVGen~\cite{swerdlow2024streetview} & \textcolor{ForestGreen}{\usym{2713}} & \textcolor{red}{\usym{2717}} & \textcolor{red}{\usym{2717}} & 25.54 & - \\
BEVControl~\cite{yang2023bevcontrol} & \textcolor{ForestGreen}{\usym{2713}} & \textcolor{red}{\usym{2717}} & \textcolor{red}{\usym{2717}} & 24.85 & - \\
 DriveGAN~\cite{kim2021drivegan} & \textcolor{red}{\usym{2717}} & \textcolor{ForestGreen}{\usym{2713}} & \textcolor{red}{\usym{2717}} & 73.40 & 502.30 \\
DriveDreamer~\cite{wang2023drivedreamer} & \textcolor{red}{\usym{2717}}  & \textcolor{ForestGreen}{\usym{2713}} & \textcolor{red}{\usym{2717}} & 52.60 & 452.00 \\
Vista~\cite{gao2024vista} & \textcolor{red}{\usym{2717}} & \textcolor{ForestGreen}{\usym{2713}} & \textcolor{red}{\usym{2717}} & 6.90 & 89.40 \\ \midrule

WoVoGen~\cite{lu2023wovogen} & \textcolor{ForestGreen}{\usym{2713}} & \textcolor{ForestGreen}{\usym{2713}} & \textcolor{red}{\usym{2717}} & 27.60 & 417.70 \\
Panacea~\cite{wen2023panacea} & \textcolor{ForestGreen}{\usym{2713}} & \textcolor{ForestGreen}{\usym{2713}} & \textcolor{red}{\usym{2717}} & 16.96 & 139.00\\
MagicDrive~\cite{gao2023magicdrive} & \textcolor{ForestGreen}{\usym{2713}} & \textcolor{ForestGreen}{\usym{2713}} & \textcolor{red}{\usym{2717}} & 16.20 & 217.94 \\ 
MagicDriveDiT~\cite{gao2024magicdrivedit} & \textcolor{ForestGreen}{\usym{2713}} & \textcolor{ForestGreen}{\usym{2713}} & \textcolor{red}{\usym{2717}} & 20.91 & 94.84 \\

Drive-WM~\cite{wang2023drivewm} & \textcolor{ForestGreen}{\usym{2713}} & \textcolor{ForestGreen}{\usym{2713}} & \textcolor{red}{\usym{2717}} &  15.80  &  122.70  \\
Vista$^*$~\cite{gao2024vista} & \textcolor{ForestGreen}{\usym{2713}} & \textcolor{ForestGreen}{\usym{2713}} & \textcolor{red}{\usym{2717}} & 13.97 & 112.65  \\ 
UniScene~\cite{li2024uniscene} & \textcolor{ForestGreen}{\usym{2713}} & \textcolor{ForestGreen}{\usym{2713}} & \textcolor{red}{\usym{2717}} &  \textbf{6.45}  &  71.94  \\ \midrule
\rowcolor{gray!10}DiST-T (Ours-D) & \textcolor{ForestGreen}{\usym{2713}}&  \textcolor{ForestGreen}{\usym{2713}} & \textcolor{ForestGreen}{\usym{2713}} & 7.40 & \underline{25.55 } \\ 
\rowcolor{gray!10}DiST-T (Ours) & \textcolor{ForestGreen}{\usym{2713}}&  \textcolor{ForestGreen}{\usym{2713}} & \textcolor{ForestGreen}{\usym{2713}} & \underline{6.83} & \textbf{22.67} \\ 
\bottomrule
\end{tabular}
 }
 \vspace{-5pt}
\caption{\textbf{Quantitative results of RGB Video Generation.} 
Evaluation on the NuScenes validation set with FID and FVD.
Unlike baselines lacking depth modeling, DiST-T generates multi-view RGB videos with depth, achieving state-of-the-art performance. The results of the multi-view variant of Vista$^*$~\cite{gao2024vista} are reported in ~\cite{li2024uniscene}.
}
\vspace{-10pt}
\label{tab_video}
\end{center}
\end{table}

\begin{table}[!t]
\begin{center}
\scriptsize
\vspace{-0pt}
\renewcommand\tabcolsep{1.0pt}
\resizebox{0.99\linewidth}{!}{
\begin{tabular}{l|cc|cccc|ccc}
\toprule 
\multicolumn{1}{l|}{\multirow{2}{*}{Method}} & \multicolumn{2}{c|}{Detection $\uparrow$} & \multicolumn{4}{c|}{BEV Segmentation $\uparrow$} & \multicolumn{3}{c}{L2 $\downarrow$}  \\
\cmidrule(lr){2-3} \cmidrule(lr){4-7} \cmidrule(lr){8-10} 
 & ~ NDS ~ & ~ mAP ~ & ~ Lan.~ & ~ Dri. ~ & ~ Div. ~ & ~ Cro. ~ & ~ 1.0 ~ & ~ 2.0 ~ &  ~ 3.0 ~  \\
\midrule
Ori. GT  & 49.85 & 37.98 & 31.31 & 69.14 & 25.93 & 14.36 & 0.51 & 0.98 & 1.65  \\
\midrule
MD~\cite{gao2023magicdrive}  & 28.36 & 12.92 & 21.95 & 51.46 & 17.10 & 5.25 & 0.57 & 1.14 & 1.95  \\
DA~\cite{yang2024drivearena}  & \underline{30.03} & \textbf{16.06} & \underline{26.14} & \underline{59.37} & \underline{20.79} & \underline{8.92} & \textbf{0.56} & \textbf{1.10} & \textbf{1.89}   \\
\rowcolor{gray!10}Ours  & \textbf{32.44} & \underline{15.63} & \textbf{26.80} & \textbf{60.32} & \textbf{21.69} & \textbf{10.99} & \textbf{0.56} & \underline{1.11} & \underline{1.91} \\
\bottomrule
\end{tabular}
}
\vspace{-5pt}
\caption{\textbf{Quantitative evaluation of video generation for perception and planning}. We use UniAD~\cite{hu2023planning} to assess detection, BEV segmentation, and L2 planning errors on generated videos compared to original nuScenes data~\cite{caesar2020nuscenes}. Our method outperforms baselines in perception and segmentation while maintaining competitive planning accuracy, showing the effectiveness of DiST-4D for various autonomous driving tasks.}
\vspace{-15pt}
\label{tab_planning}
\end{center}
\end{table}

\paragraph{Evaluation Metric.}
We evaluate the quality of the generated RGB videos using the FID \cite{Seitzer2020FID} for images and FVD for videos.
The generated RGB videos have a resolution of $424 \times 800 \times 6$ and a total of $17$ frames.

Since DiST-T conditions RGB-D video generation on a reference image, the first frame of the reference image is excluded from evaluations.
For depth sequence generation, as no prior method jointly generates RGB-D videos in driving scenarios, we compare the depth generation performance with SOTA surround depth estimation methods \cite{wei2023surrounddepth,zou2024m2depth} using common metrics including Absolute Relative Error (Abs.Rel.), Root Mean Squared Error (RMSE) and Threshold Accuracy ($\delta$) as doing in~\cite{wei2023surrounddepth}.

\begin{table}[!t]
\centering
\resizebox{0.99\linewidth}{!}{
\begin{tabular}{l|cccccc}
\toprule
\textbf{Method} &  Abs. Rel. $\downarrow$  & RMSE $\downarrow$ & $\delta < 1.25$ $\uparrow$ & $\delta < 1.25^2$ $\uparrow$  \\
\midrule
\multicolumn{5}{c}{\textit{Single Frame LiDAR GT}} \\
\midrule
SD~\cite{wei2023surrounddepth} &  0.27 / 0.28  & 7.29 / 7.47 & 0.68 / 0.66 & 0.86 / 0.84 \\
M$^2$Depth ~\cite{zou2024m2depth}        &  0.24 / \textbf{0.26} & 6.82 / \textbf{6.90} & 0.72 / \textbf{0.73} & 0.88 / 0.87  \\
\rowcolor{gray!10}DiST-T (Ours-D) & \textbf{0.21} / 0.27 &   \textbf{6.79} / 7.06 &  \textbf{0.77} / \textbf{0.73} & \textbf{0.89} / \textbf{0.88} \\
\rowcolor{gray!10}DiST-T (Ours)                & 0.25 / 0.39 & 7.39 / 7.75 & 0.70 / 0.58 & 0.87 / 0.81 \\
\midrule
\multicolumn{5}{c}{\textit{Multiple Frame LiDAR GT}} \\
\midrule
SD~\cite{wei2023surrounddepth} & 0.27 / 0.28 & 6.50 / 6.59 & 0.67 / 0.63 & 0.87 / 0.85 \\
M$^2$Depth  ~\cite{zou2024m2depth}         &  0.25 / 0.26 & 6.02 / 6.16 & 0.72 / 0.72 & 0.89 / 0.88 \\
\rowcolor{gray!10}DiST-T (Ours-D)               & \textbf{0.20} / \textbf{0.25} & \textbf{5.33} / \textbf{5.58} & \textbf{0.79} / \textbf{0.75} & \textbf{0.92} / \textbf{0.91} \\
\rowcolor{gray!10}DiST-T (Ours)               & 0.24 / 0.35 & 6.24 / 6.89 & 0.71 / 0.61 & 0.88 / 0.83 \\
\bottomrule
\end{tabular}}
\vspace{-5pt}
\caption{\textbf{Quantitative evaluation of generated depth.} We compare DiST-T with state-of-the-art surround depth estimation methods~\cite{wei2023surrounddepth, zou2024m2depth} on the nuScenes dataset~\cite{caesar2020nuscenes}. Results are reported \textit{with} / \textit{without} median scaling for both single-frame and multi-frame LiDAR ground truth. 
}
\label{tab:depth_generation_results}
\end{table}

\begin{figure}[!t]
    \centering
    \includegraphics[width=1.0\linewidth]{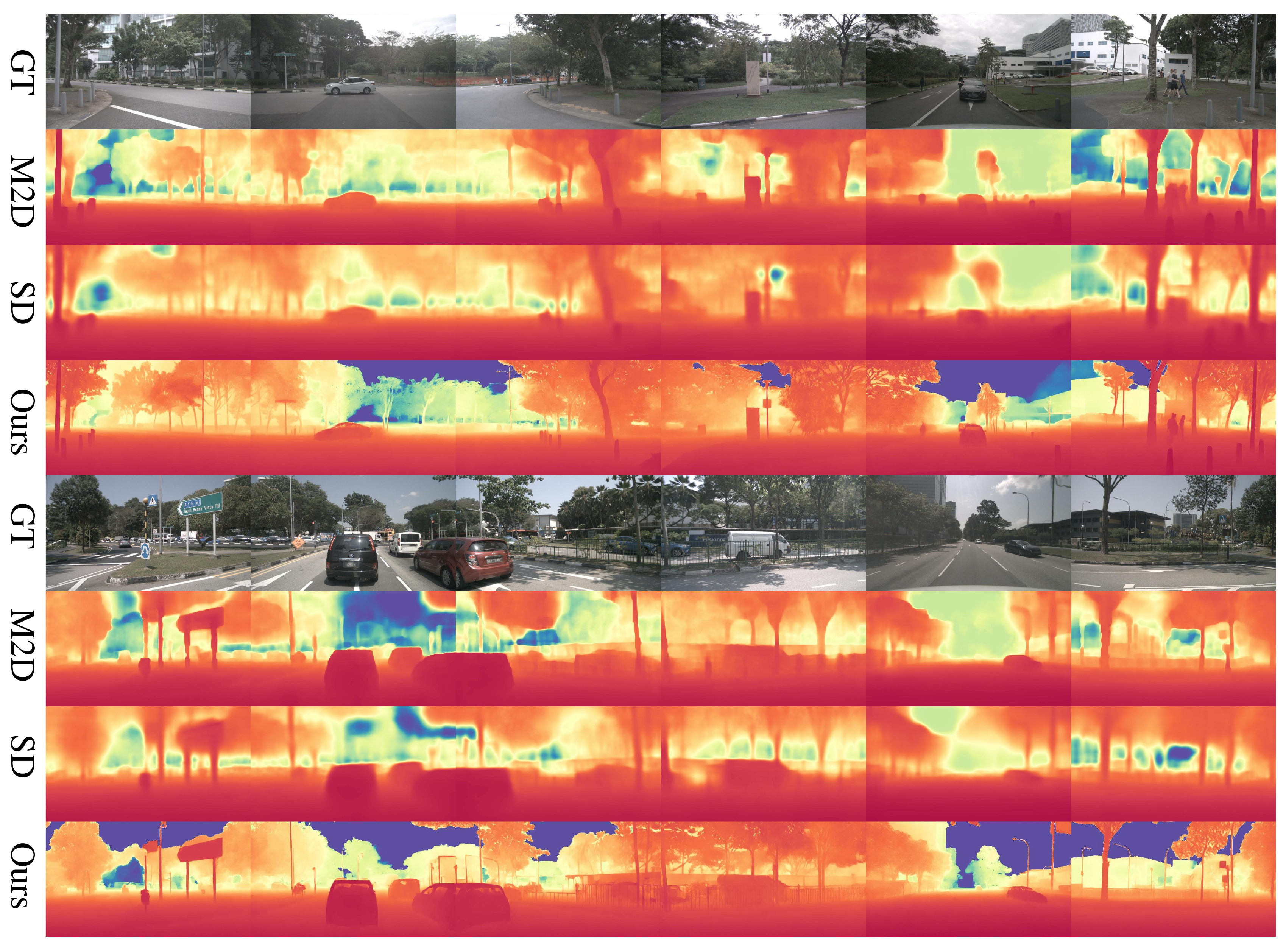}
    \vspace{-18pt}
    \caption{\textbf{Visualization results of generated surround depth.} Compared to M$^2$Depth~\cite{zou2024m2depth} (M2D) and SurroundDepth~\cite{wei2023surrounddepth} (SD), DiST-T produces more fine-grained depth with enhanced details.}
    \label{DiST_T_depth_compare}
    \vspace{-8pt}
\end{figure}

\paragraph{RGB Video Generation Quality.}
The quantitative comparison of video generation is illustrated in Tab.~\ref{tab_video}.
DiST-T achieves a remarkable improvement in the FVD metric compared to previous models, indicating high fidelity in the generated videos. In terms of FID, DiST-T achieves 6.83, which is close to the previous SOTA model~\cite{li2024uniscene}.
We also modify DiST-T to take multi-modal (RGB+depth) reference images as input for comparison, denoted as Ours-D, where the reference depth is obtained from our depth curation pipeline.
Tab.~\ref{tab_video} indicates that incorporating reference depth has certain side effects on RGB forecasting.

\vspace{-10pt}
\paragraph{Downstream Evaluation.}
In addition to evaluating the visual quality of the generated images and videos, we assess their effectiveness in downstream tasks, such as perception and open-loop planning using UniAD~\cite{hu2023planning}. The quantitative results in Tab.~\ref{tab_planning} show that the high-quality videos produced by DiST-T achieve performance comparable to the original ground truth data, highlighting the potential of synthetic data for downstream applications.

\begin{figure}[!t]
    \centering
    \includegraphics[width=1.0\linewidth]{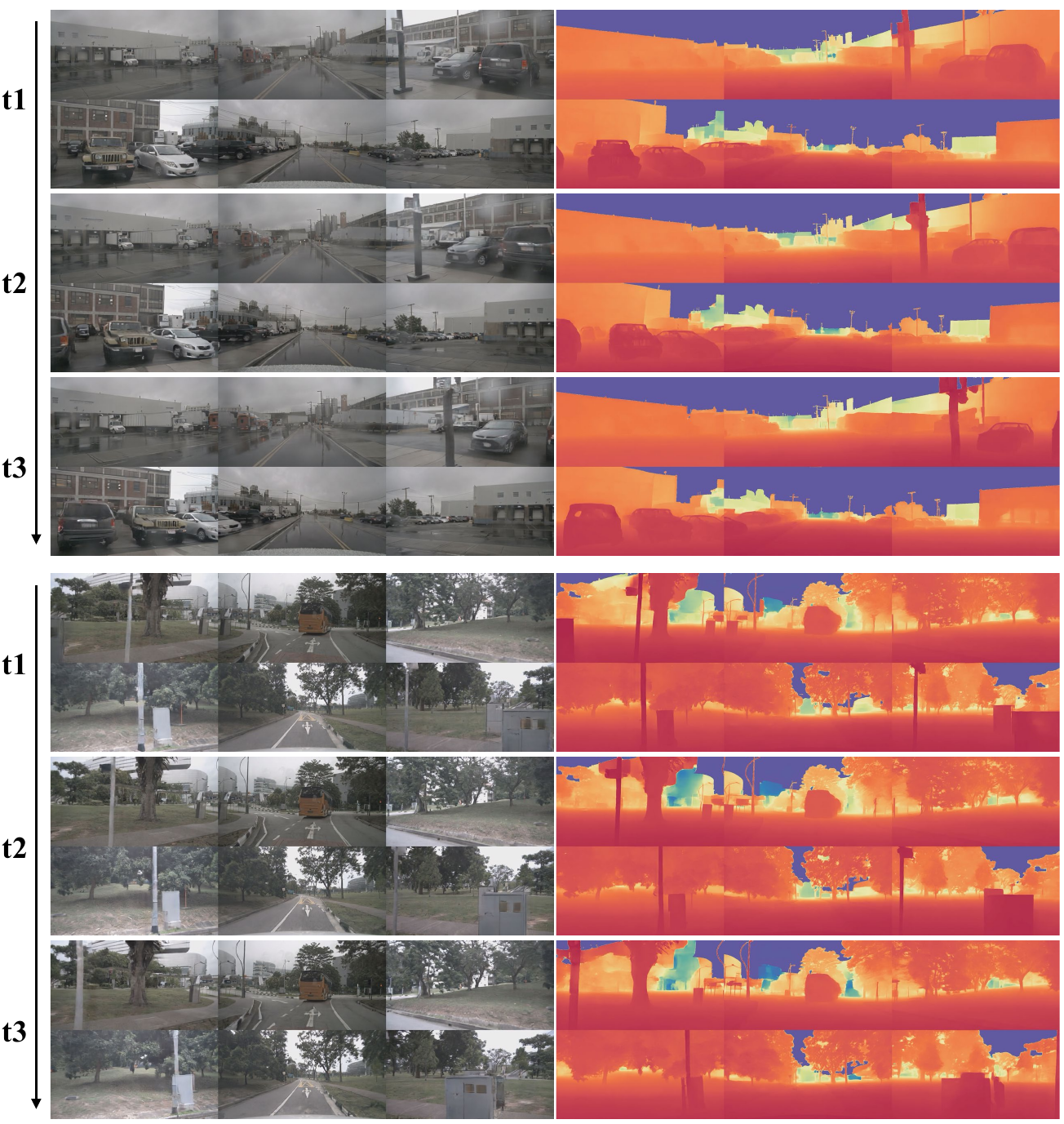}
    \vspace{-15pt}
    \caption{\textbf{Visualization results of DiST-T.} DiST-T produces high-fidelity multi-view RGB videos (left) and fine-grained depth sequences (right) over time steps. 
    }
    \label{DiST_T_mm_vis}
    \vspace{-5pt}
\end{figure}

\paragraph{Depth Generation Quality.}
For depth generation, we report both the results of single-frame LiDAR GT~\cite{wei2023surrounddepth} and our multi-frame LiDAR GT, which are reported \textit{with} / \textit{without} median scaling.
It is worth noting that our model is different from depth estimation models~\cite{zou2024m2depth,wei2023surrounddepth}, which take GT image as input to predict the corresponding depth, our method simultaneously generates future RGB frames and their corresponding depth sequences with only one frame of reference images.
As shown in Tab.~\ref{tab:depth_generation_results}, the generated future depths (Ours) achieve comparable performance to~\cite{zou2024m2depth, wei2023surrounddepth}.
When the reference depths are available for the modified DiST-T (Ours-D), our model surpasses existing models on both single-frame and multi-frame depth GT.

The qualitative results of the RGB-D sequence generated by DiST-T are shown in
Fig.~\ref{DiST_T_mm_vis}.
Even in complex driving scenarios, DiST-T can generate high-quality RGB videos along with fine-grained depth maps.
Additional visualizations of DiST-T are provided in the supplementary materials.

\begin{figure*}[!t]
    \centering
    \vspace{-10pt}
    \includegraphics[width=0.95\linewidth]{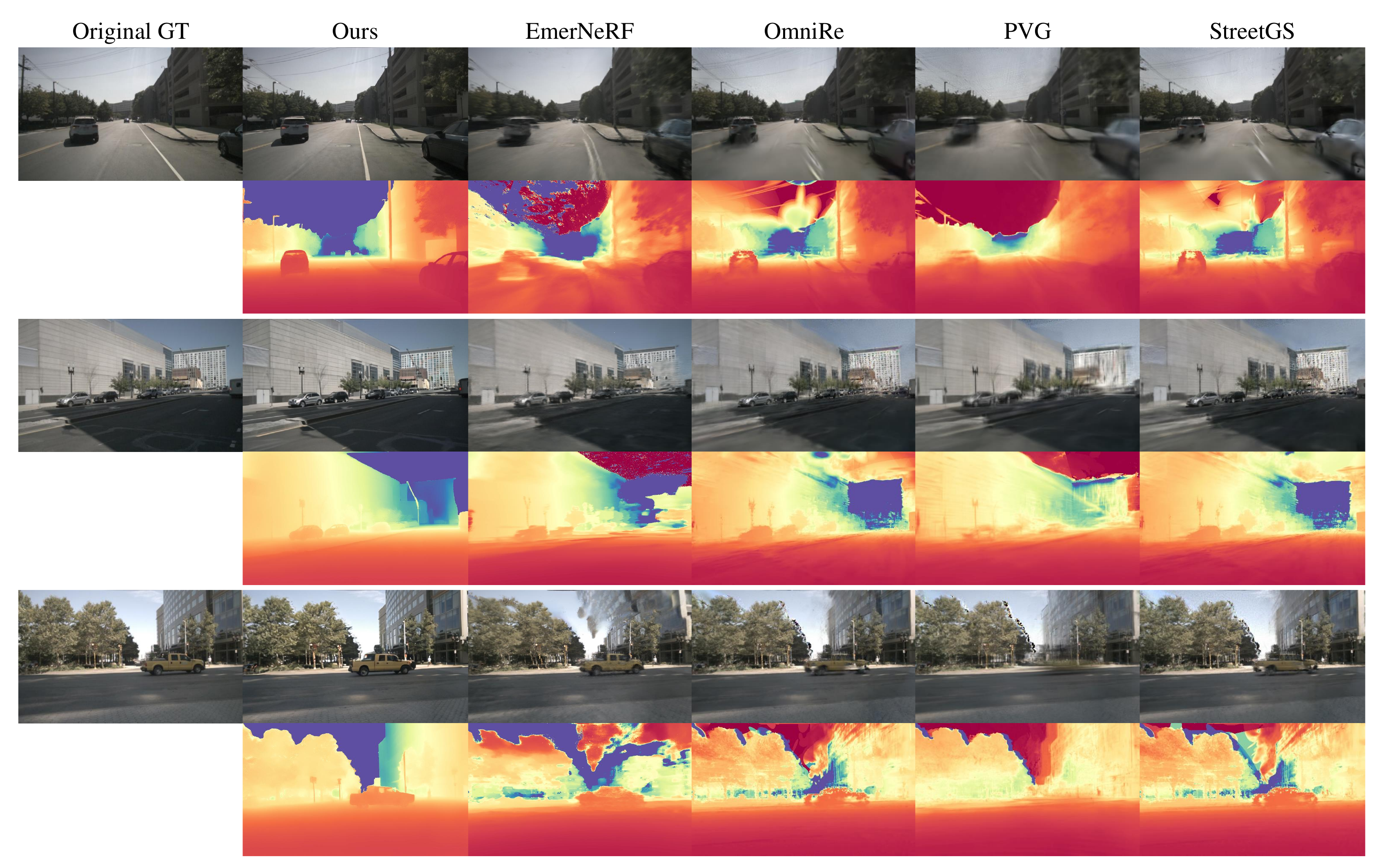}
    \vspace{-10pt}
    \caption{\textbf{Qualitative comparison on RGB-D spatial NVS}. Comparison between DiST-S and state-of-the-art reconstruction-based methods~\cite{yang2023emernerf, chen2024omnire, chen2023PVG, yan2024street} on spatial NVS. DiST-S preserves fine-grained scene geometry and reduces artifacts in shifted viewpoints.}
    \label{DiST_S_compare_vis}
    \vspace{-10pt}
\end{figure*}

\subsection{Spatial Novel View Synthesis}
\paragraph{Evaluation Metric.}

Following the evaluation methodology introduced in~\cite{wang2024freevs}, we primarily assess the performance of DiST-S on novel trajectories with FID and FVD. We apply different lateral offsets $\tau \in \{\pm 1m, \pm 2m, \pm 4m \}$ to the camera viewpoints and calculate the FID and FVD metrics between the synthesized RGB images of the novel trajectory and the ground truth images of the original trajectory. 

\begin{table}[!t]
\begin{center}
\scriptsize
\vspace{-0pt}
\renewcommand\tabcolsep{1.0pt}
\resizebox{1.0\linewidth}{!}{%
\begin{tabular}{l|cc|cc|cc}
\toprule 
\multicolumn{1}{l|}{\multirow{2}{*}{Method}} & \multicolumn{2}{c|}{Shift $\pm$1m} & \multicolumn{2}{c|}{Shift $\pm$2m} & \multicolumn{2}{c}{Shift $\pm$4m} \\
\cmidrule(lr){2-3} \cmidrule(lr){4-5} \cmidrule(lr){6-7}
 & ~ FID $\downarrow$ ~ & ~FVD $\downarrow$ ~ & ~ FID $\downarrow$ ~ & ~ FVD $\downarrow$ ~ & ~ FID $\downarrow$ ~ & ~ FVD $\downarrow$ ~ \\
\midrule
PVG~\cite{chen2023PVG}      &  48.15 & 246.74  & 60.44 & 356.23 & 84.50 & 501.16  \\
EmerNeRF~\cite{yang2023emernerf}   &  37.57  &  171.47  & 52.03  & 294.55 & 76.11 &  497.85 \\ 
StreetGaussian~\cite{yan2024street} & 32.12  & 153.45  &  43.24 & 256.91 & 67.44 & 429.98 \\
OmniRe~\cite{chen2024omnire}   & 31.48  & 152.01  & 43.31  & 254.52 & 67.36 & 428.20 \\
FreeVS$^*$~\cite{wang2024freevs}   &  51.26  &  431.99  &  62.04  &  497.37  &  77.14  &  556.14 \\
\midrule
\rowcolor{gray!10}DiST-4D (Ours)  &  20.64 &  130.98 &  25.08  &  156.60  &  33.56  &  189.04  \\
\rowcolor{gray!10}DiST-4D (+SCC)  &  16.40 &  112.86 &  19.50  &  124.97  &  25.16  &   146.56 \\
\rowcolor{gray!10}DiST-S (Ours)   &  12.00 &  55.80 &  15.72  &  89.54  &  22.54  &  137.03  \\
\rowcolor{gray!10}DiST-S (+SCC)   &  \textbf{10.12} &  \textbf{45.14} &  \textbf{12.97}  &  \textbf{68.80}  &  \textbf{17.57}  &  \textbf{105.29}  \\
\bottomrule
\end{tabular}
}
\vspace{-5pt}
\caption{\textbf{Quantitative comparison of novel view synthesis}. We evaluate FID and FVD across different viewpoint shifts (±1m, ±2m, ±4m). FreeVS{$^*$} denotes we retrain FreeVS~\cite{wang2024freevs} using the official code on the nuScenes dataset~\cite{caesar2020nuscenes}. DiST-4D uses generated RGB-D videos from DiST-T, while DiST-S leverages real videos with preprocessed metric depth. DiST-S (+SCC) achieves the best performance, demonstrating the effectiveness of self-supervised cycle consistency in improving novel view synthesis quality.
}
\vspace{-12pt}
\label{tab_NVS}
\end{center}
\end{table}

\begin{table}[!t]
  \centering
  \resizebox{0.99\linewidth}{!}{
    \begin{tabular}{l|ccc|cc}
    \toprule
    & Depth & Valid Mask & Data Aug. & FID-1m $\downarrow$& FID-2m $\downarrow$ \\
    \midrule
    (a)  &   -    & -     & \usym{2713}     & 26.33 & 33.54 \\
    (b) & \usym{2713}     &   -   & \usym{2713}     & 30.31 & 32.80 \\
    (c) & \usym{2713}    & \usym{2713}     &  -  & 26.19 & 29.69 \\
    \midrule
    DiST-S &  \usym{2713}  & \usym{2713}  & \usym{2713}  & \textbf{25.51} & \textbf{27.75} \\
    \bottomrule
    \end{tabular}
    }%
    \vspace{-5pt}
    \caption{\textbf{Ablation study of DiST-S}. We analyze the impact of depth modality, valid mask usage, and data augmentation on synthesis quality, reporting FID at 1m and 2m shifts. Results are evaluated on the front camera of a selected nuScenes validation subset~\cite{caesar2020nuscenes}.}
    \vspace{-10pt}
  \label{tab:NVS_ab}%
\end{table}%

\paragraph{Benchmark Evaluation.}
In Tab.~\ref{tab_NVS}, we compare the NVS results of DiST-S with existing methods under shifted novel views. The quantitative results demonstrate that our method achieves substantial improvements in both FID and FVD metrics.
When using conditions generated from real images as inputs, DiST-S achieves several times better performance in these metrics, highlighting its strong capability in NVS tasks based on real data. As illustrated in Fig. \ref{DiST_S_compare_vis}, the proposed DiST-S significantly outperforms previous reconstruction-based models in synthesizing novel spatial viewpoints for both RGB and depth, which is nearly free from image degradation and artifacts.
Even when using conditions from generated videos of DiST-T (denoted as DiST-4D), our model achieves significantly lower FID and FVD scores than other methods.

\begin{figure}[!t]
    \centering
    \includegraphics[width=1.0\linewidth]{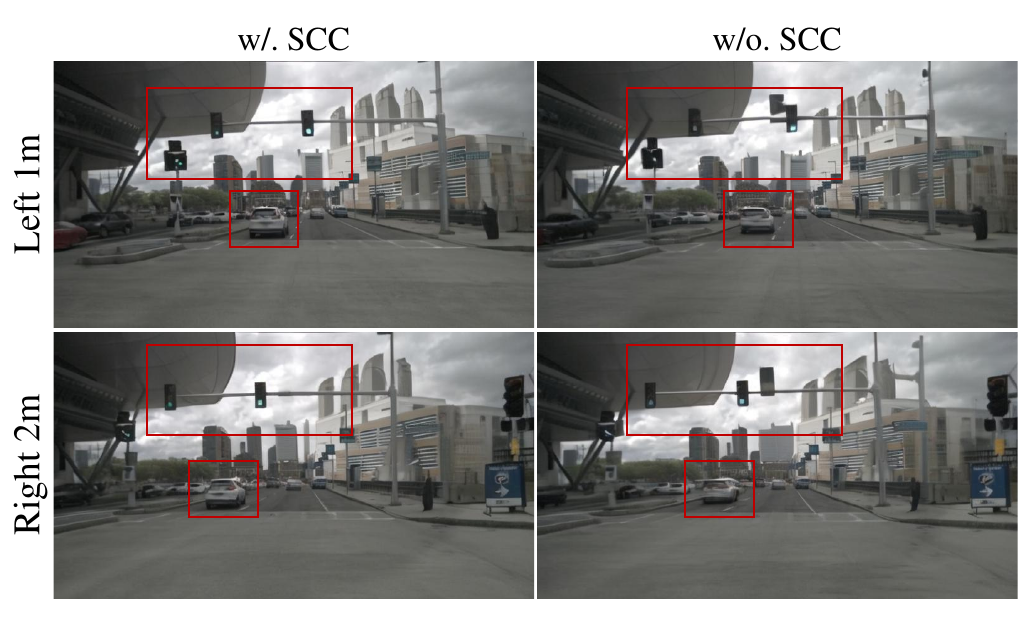}
    \vspace{-18pt}
    \caption{\textbf{Qualitative ablation study on the SCC strategy}. We compare novel view synthesis results with and without self-supervised cycle consistency (SCC) under different viewpoint shifts (left 1m and right 2m). Red boxes highlight improved object fidelity and scene consistency when SCC is applied, particularly in fine-grained structures like traffic lights and vehicles.}
    \vspace{-16pt}
    \label{ab_scc_vis}
\end{figure}

\subsection{Ablation Study}

We conduct ablation studies for DiST-S in Tab.~\ref{tab:NVS_ab}, reporting FID and FVD results for novel trajectory synthesis with lane shifting. First, we modify DiST-S into an RGB-only spatial NVS model, denoted as (a). By comparing (a) and naive multimodel DiST-S (b), we find that multi-modal shows a slight decrease in FID. We observe that \textit{Data Augmentation} with $n=\pm4$ frame projection and point cloud downsampling outperforms (c) in generation quality by 2.5\%. The results in (b) demonstrate that incorporating the \textit{Valid Mask} significantly enhances DiST-S's capability, reducing the FID score by 15.8\%.

Additionally, we perform an ablation study on the proposed self-supervised cycle consistency (SCC) strategy. As shown in Fig.~\ref{ab_scc_vis} and Tab.~\ref{tab_NVS}, SCC leads to substantial improvements in dynamic vehicle representation and fine object details, significantly boosting the quality of novel view synthesis, with the FID metric also reduced by 20\%.

\section{Conclusion and Future Work}
In this paper, we introduce DiST-4D, a novel framework for feed-forward 4D driving scene generation. With a disentangled spatiotemporal paradigm that bridges temporal forecasting and spatial novel view synthesis with dense metric depth, DiST-4D is capable of generating RGB-D 4D scenarios without per-scene optimization. Extensive experiments show the DiST-4D achieves state-of-the art performance in both temporal and spatial generation. 

\vspace{-10pt}
\paragraph{Limitations and future work}

DiST-4D leverages metric depth to connect future temporal predictions with novel spatial viewpoint synthesis. However, its performance is still influenced by the quality of the pseudo metric depth ground truth, especially when it comes to distant buildings. This underscores the need for further research on acquiring high-quality metric depth in outdoor environments. Additionally, expanding DiST-4D's application to areas such as embodied intelligence and robotics offers a promising avenue for future development.

\newpage
{
    \small
    \bibliographystyle{ieeenat_fullname}
    \bibliography{main}
}

\newpage
\appendix
\maketitle
\twocolumn[{
\centering
\vspace{20pt}
\section*{\Large \centering Supplementary Material}
 \vspace{30pt}
 }]

\section{Metric Depth Curation}
The proposed metric depth curation pipeline relies on LiDAR point clouds and visual
reconstruction results to generate fine-grained depth maps.
Due to the high sparsity of LiDAR point clouds in the nuScenes~\cite{caesar2020nuscenes} dataset, we aggregate multi-frame LiDAR data using a window of three frames. For the MVS reconstruction network, we set the maximum depth to 100 meters, and the final aggregated point cloud undergoes voxel downsampling with a resolution of 0.1 meters to reduce redundancy.

Since MVS static point clouds contain significant noise, we filter out ground-level noise by retaining only points located above the ego vehicle’s LiDAR. When merging LiDAR and MVS points, we prioritize LiDAR data and utilize MVS points only in regions where LiDAR coverage is unavailable. We apply nearest-neighbor interpolation to obtain an initial dense metric depth prompt. Finally, we utilize a generative depth completion network to obtain a dense and accurate metric depth map, while a semantic segmentation network is applied to identify the sky region and assign it a depth of 100 meters, ensuring consistency in depth representation.

\begin{figure*}[!htbp]
    \centering
    \includegraphics[width=1\linewidth]{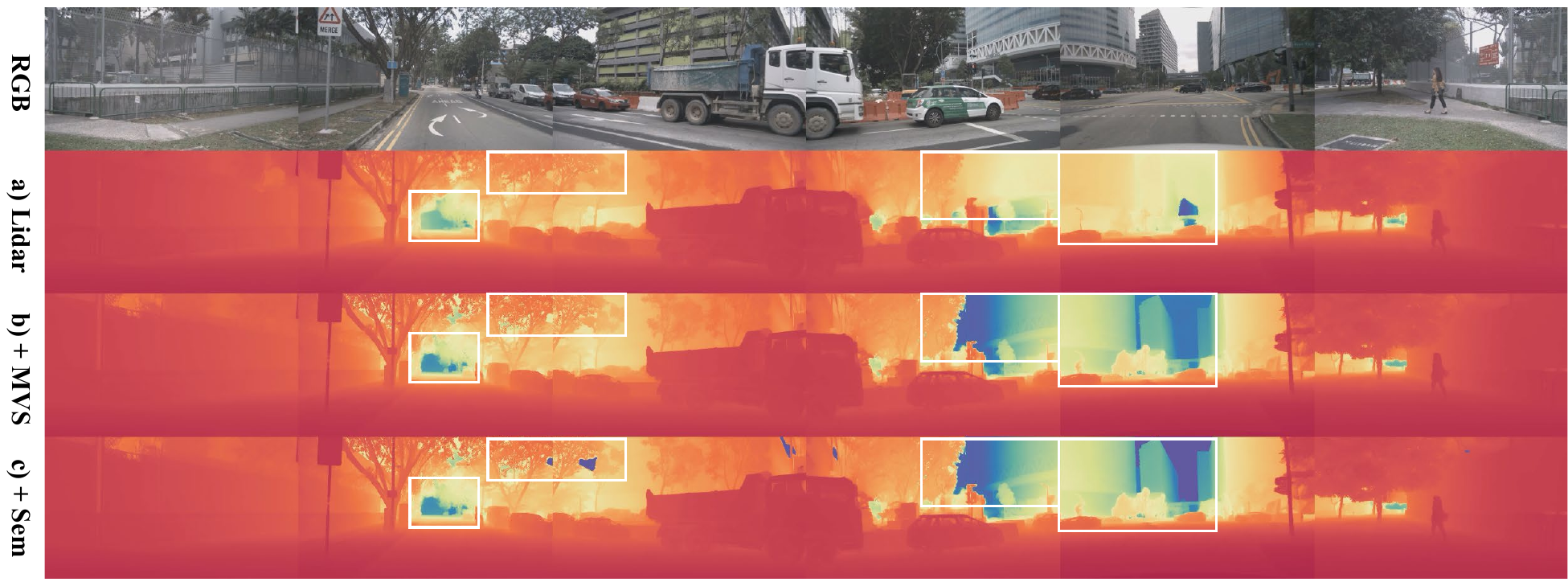}
    \caption{The comparison of the dense metric depth completion performance on a) only multi-frame LiDAR point cloud, b) add static scene point cloud with MVS, c) add sky semantic mask}
    \label{fig_DepthGen_vis_method}
\end{figure*}

The effectiveness of the proposed metric depth curation pipeline is illustrated in Fig.\ref{fig_DepthGen_vis_method}. When relying solely on LiDAR points, depth estimation errors for distant objects tend to be significant. Incorporating the static scene point cloud reconstructed via MVS substantially alleviates this issue.
Moreover, since both LiDAR and MVS-reconstructed point clouds maintain cross-camera consistency, our depth completion pipeline not only enhances fine-grained details within individual frames but also ensures high temporal and multi-view consistency across the entire scene.

We further evaluate our pseudo metric depth ground truth on the nuScenes validation set~\cite{caesar2020nuscenes}, using multi-frame LiDAR depth as the reference. As shown in Tab.\ref{tab:depth_generation_results}, our pseudo depth GT achieves higher accuracy compared to the estimated depth results from\cite{wei2023surrounddepth,zou2024m2depth}.

\begin{figure*}[!htbp]
    \centering
    \includegraphics[width=0.9\linewidth]{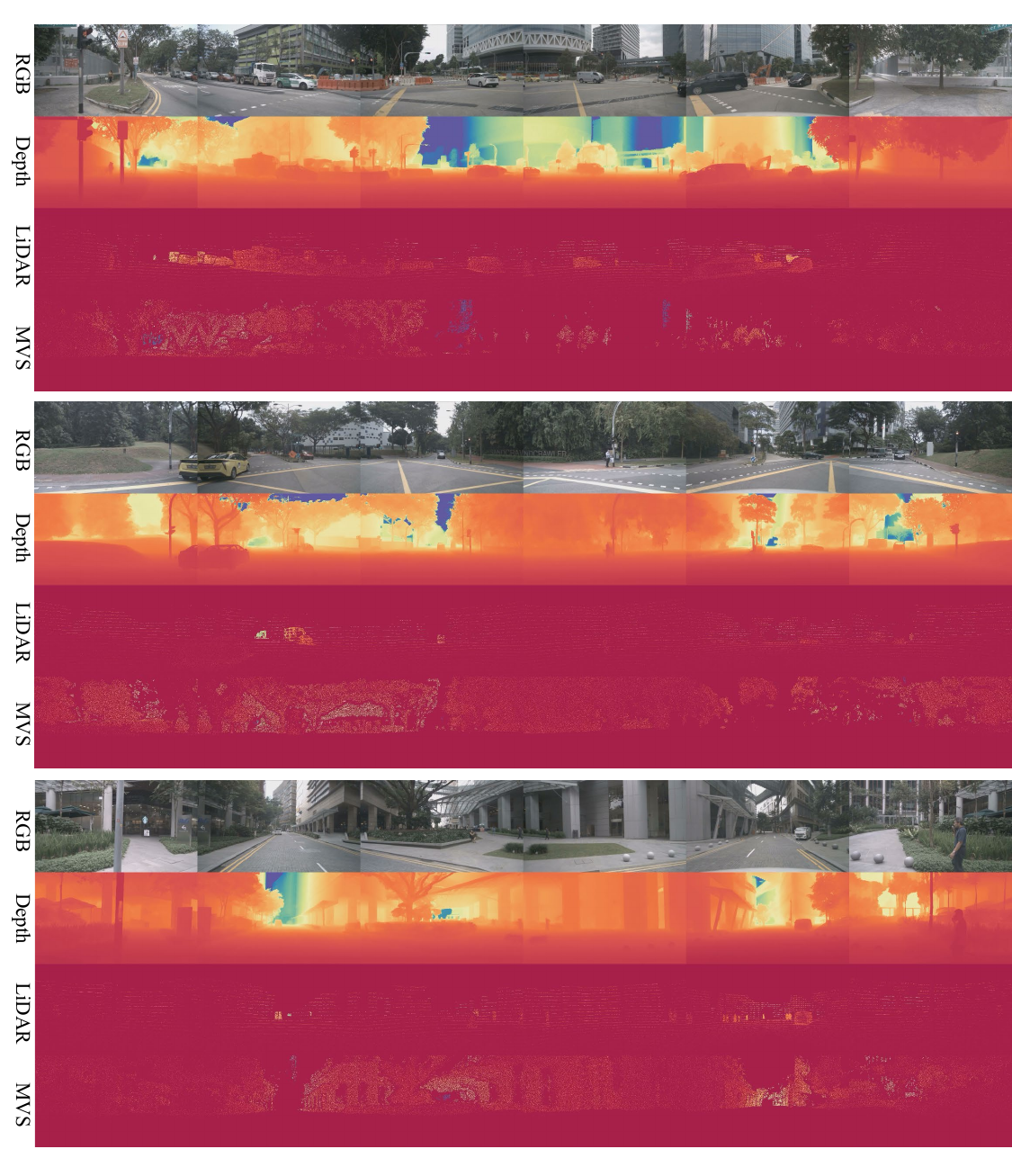}
    \caption{The visualization results of the corresponding LiDAR points and MVS points for the processed metric depth pseudo ground truth. }
    \label{fig_RGB_Depth_GT}
\end{figure*}

More results about our metric depth are in Fig.\ref{fig_RGB_Depth_GT}

\section{DiST-T}
\paragraph{Model Setup.} We use the pre-trained 3D VAE from CogVideoX \cite{yang2024cogvideox} and train the diffusion model from scratch. First, we train DiST-T for RGB video generation with a resolution of $224\times400$ for 7 days. Then we train the DiST-T for RGB-D video generation with the resolution of $224\times400$ for 3 days, followed by training at $424\times800$ for 3 days. All training phases are conducted on $8 \times$ NVIDIA H20 GPUs. The backbone of STDIT has the same layer $N=28$ and hidden size $d=1152$ following the previous work.

\begin{figure*}[!t]
    \centering
    \includegraphics[width=0.95\linewidth]{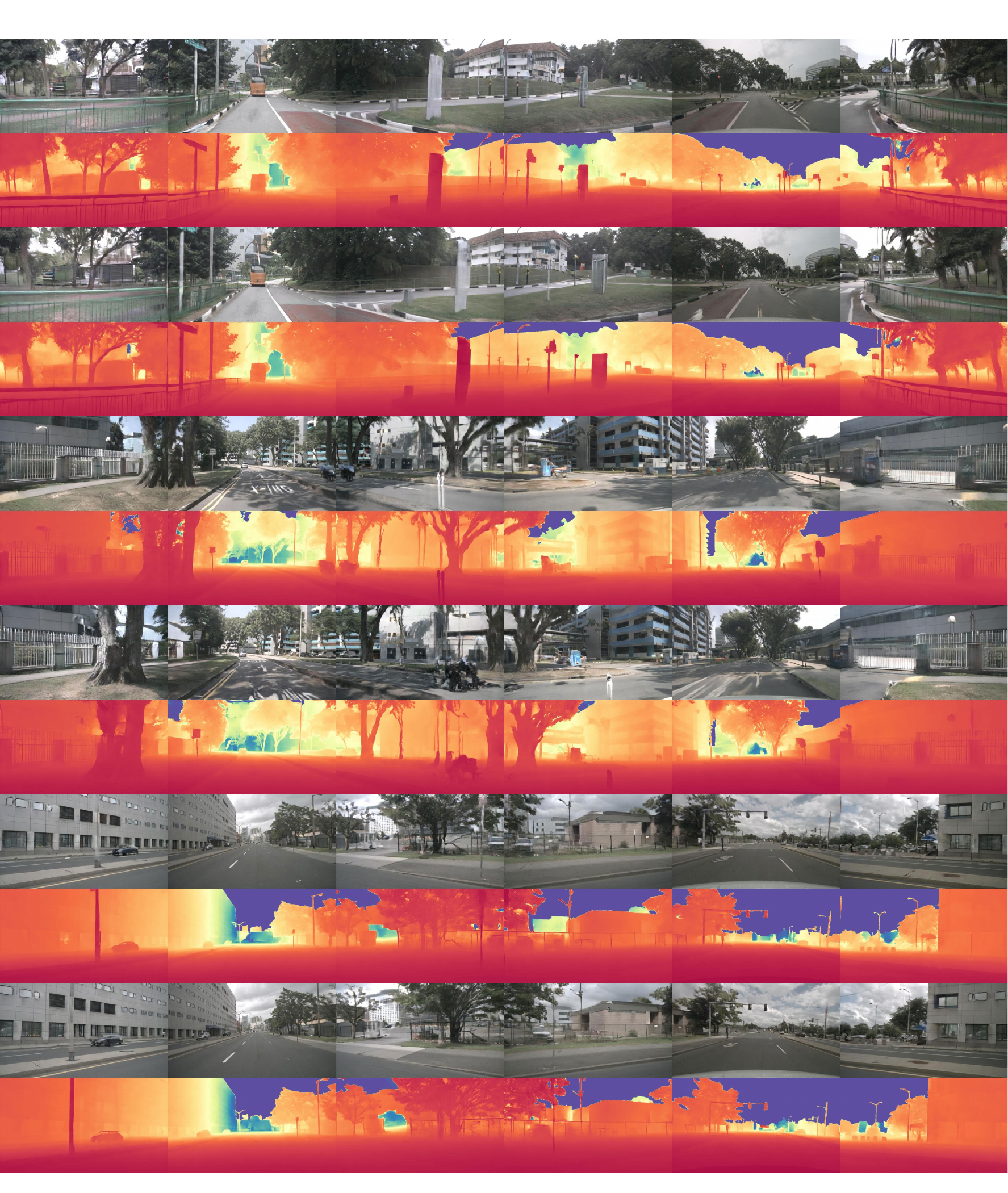}
    \caption{Additional visualizations of video generation using DiST-T. Our model can produce high-quality RGB videos along with corresponding metric depth sequences.}
    \label{fig_T1}
\end{figure*}

\begin{figure*}[!t]
    \centering
    \includegraphics[width=0.95\linewidth]{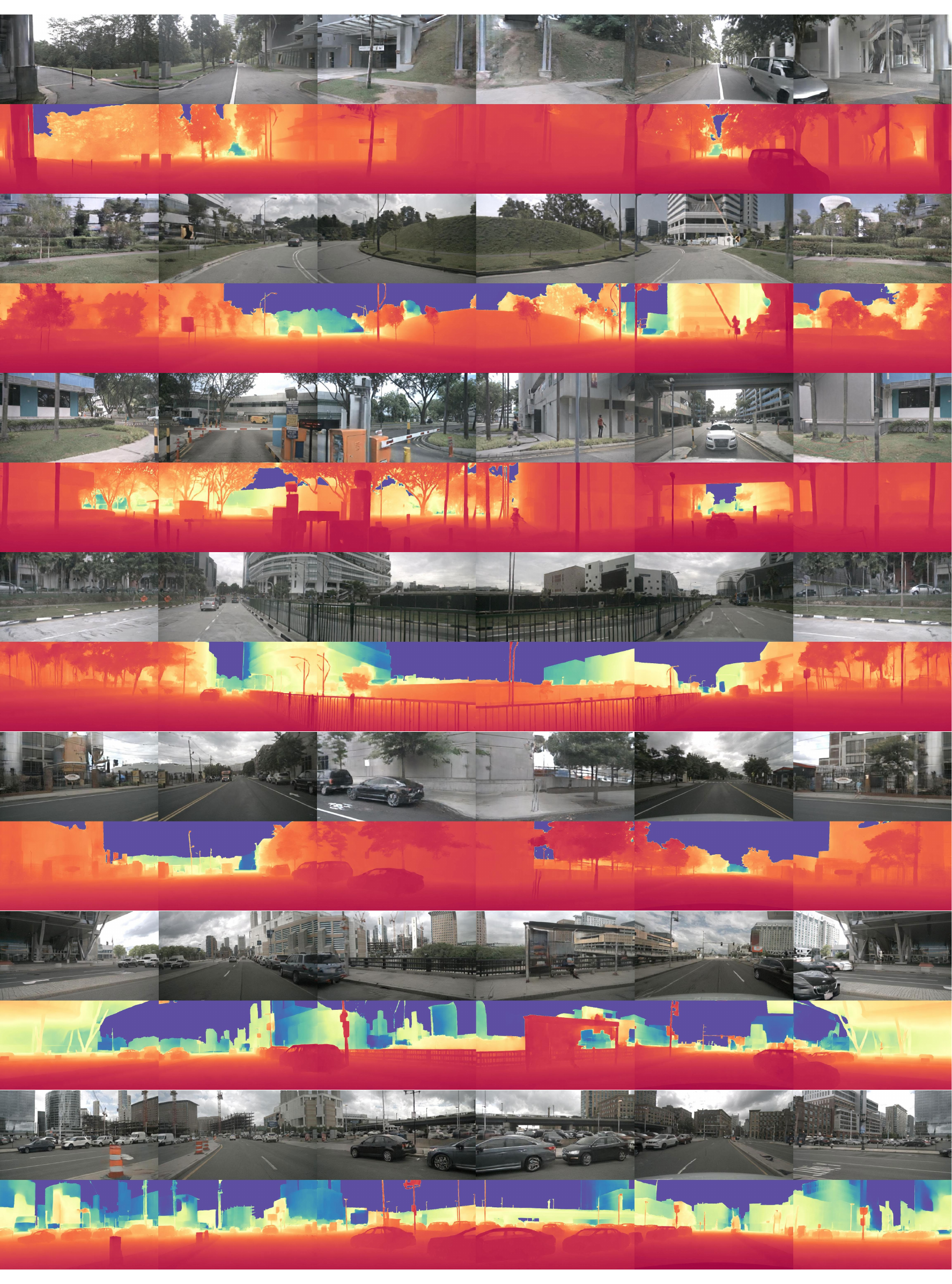}
    \caption{Additional visualizations of video generation using DiST-T. Our model can produce high-quality RGB videos along with corresponding metric depth sequences.}
    \label{fig_T4}
\end{figure*}

More visualization results about RGB-D video generation of DiST-T are provided in Fig.~\ref{fig_T1} $\sim$ Fig.~\ref{fig_T4}.

\paragraph{Loss Function}
We use simulation-free rectified flow ~\cite{lipman2022flow} and v-prediction loss~\cite{esser2024scaling}:
\begin{align}
    \mathbf{z}_{t} &= (1 - t) x_{0} + t\epsilon\\
    \mathcal{L} &= \mathbb{E}_{\epsilon\sim \mathcal{N}(0,I)}\|\mathcal{G}_{\theta} (\mathbf{z}_{t},t)-\frac{1}{1-t}(\mathbf{z}-\epsilon)\|_{2}^{2}\text{,}
\end{align}
where $t\sim\operatorname{lognorm}(0,1)$ is timestep and $\mathcal{G}_{\theta}$ indicates DiST-T network with parameters $\theta$.


\section{DiST-S}
\paragraph{Model Setup.}
The resolution of generated results is set to $448\times768$ with the video length $T=6$.
Initializing DiST-S from SVD~\cite{blattmann2023svd}, we train DiST-S for 1 day with $8\times$ NIVIDA H20 GPUs. SCC strategy is employed to full trainset for generating novel trajectories. DiST-S with the SCC strategy will require an additional day of training.

\begin{figure*}[!t]
    \centering
    \includegraphics[width=0.95\linewidth]{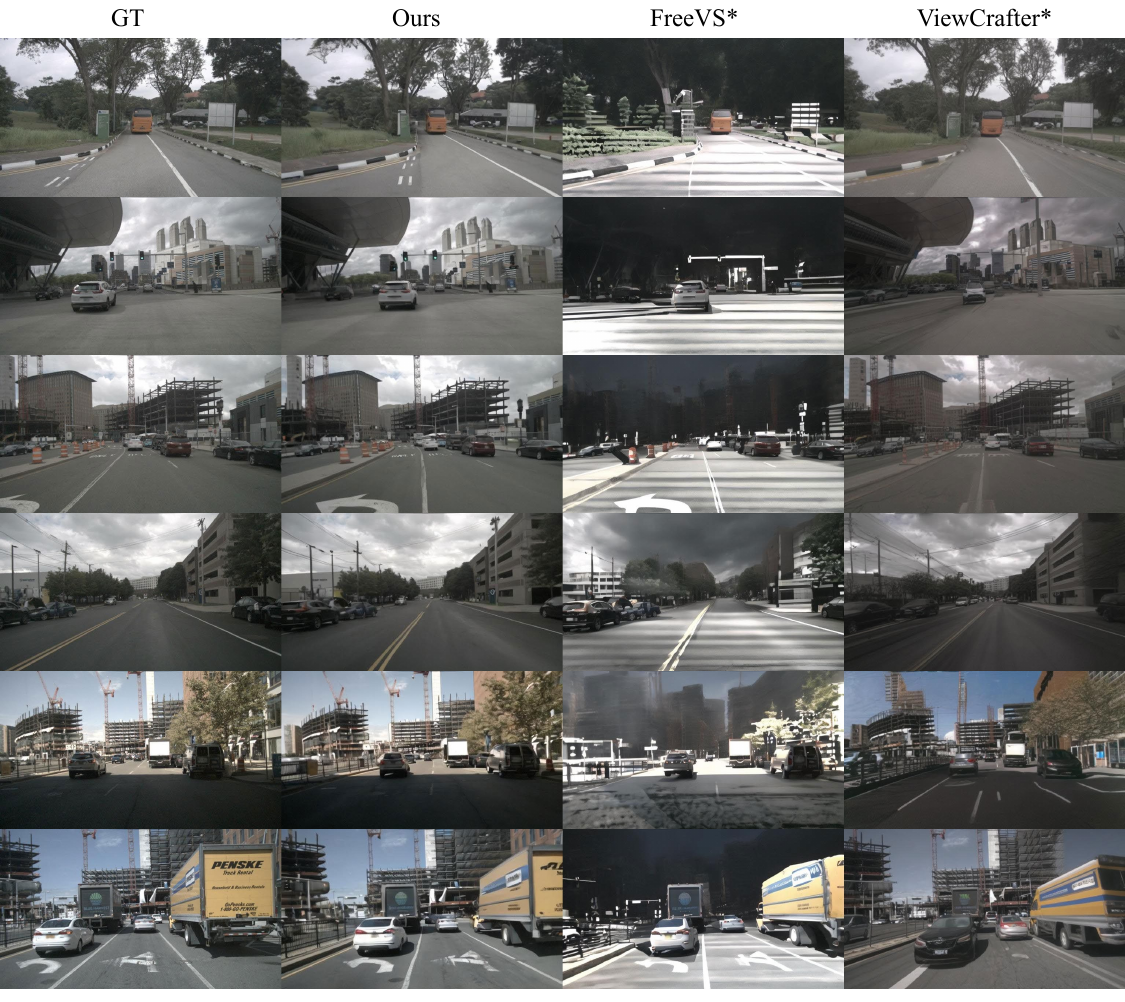}
    \caption{Comparison of FreeVS$^{*}$~\cite{wang2024freevs} and ViewCrafter$^{*}$~\cite{yu2024viewcrafter}. We present novel view synthesis results under shifted viewpoints~(shift left by 2 meters), where our method produces higher-quality images. While ViewCrafter$^{*}$ uses the same sparse conditions as ours, it struggles to adhere accurately to conditions, resulting in mismatched novel view outputs.}
    \label{VC_FreeVS}
\end{figure*}

\paragraph{More experiment results }
We qualitatively compare DiST-S with another Point2Video diffusion model, as illustrated in Fig.\ref{VC_FreeVS}. 
For FreeVS$^{*}$~\cite{wang2024freevs}, we retrain the model on the nuScenes\cite{caesar2020nuscenes} dataset using the official code and aggregated LiDAR point clouds from more frames ($n=10$). However, this method relies on accurate and dense LiDAR point clouds. Even after multi-frame aggregation, the density of LiDAR point clouds in nuScenes still falls short of that in the Waymo Open Dataset ~\cite{sun2020scalability}. Besides, LiDAR-based methods struggle to handle distant buildings. From the visual comparisons, it can be observed that FreeVS* exhibits significantly poorer performance in distant objects and the sky.

For ViewCrafter$^{*}$\cite{yu2024viewcrafter}, DUSt3R\cite{dust3r_cvpr24} is used for relative depth estimation in the official model, making control-specific locations unfeasible. Therefore, we utilized our processed pseudo-image as the conditional input and the image in the recorded trajectory as a reference image. However, as this method is specifically designed for static scenes, its effectiveness in driving scenarios is limited, and it lacks the capability to accurately synthesize novel views based on given conditions.

\begin{figure*}[!t]
    \centering
    \includegraphics[width=0.95\linewidth]{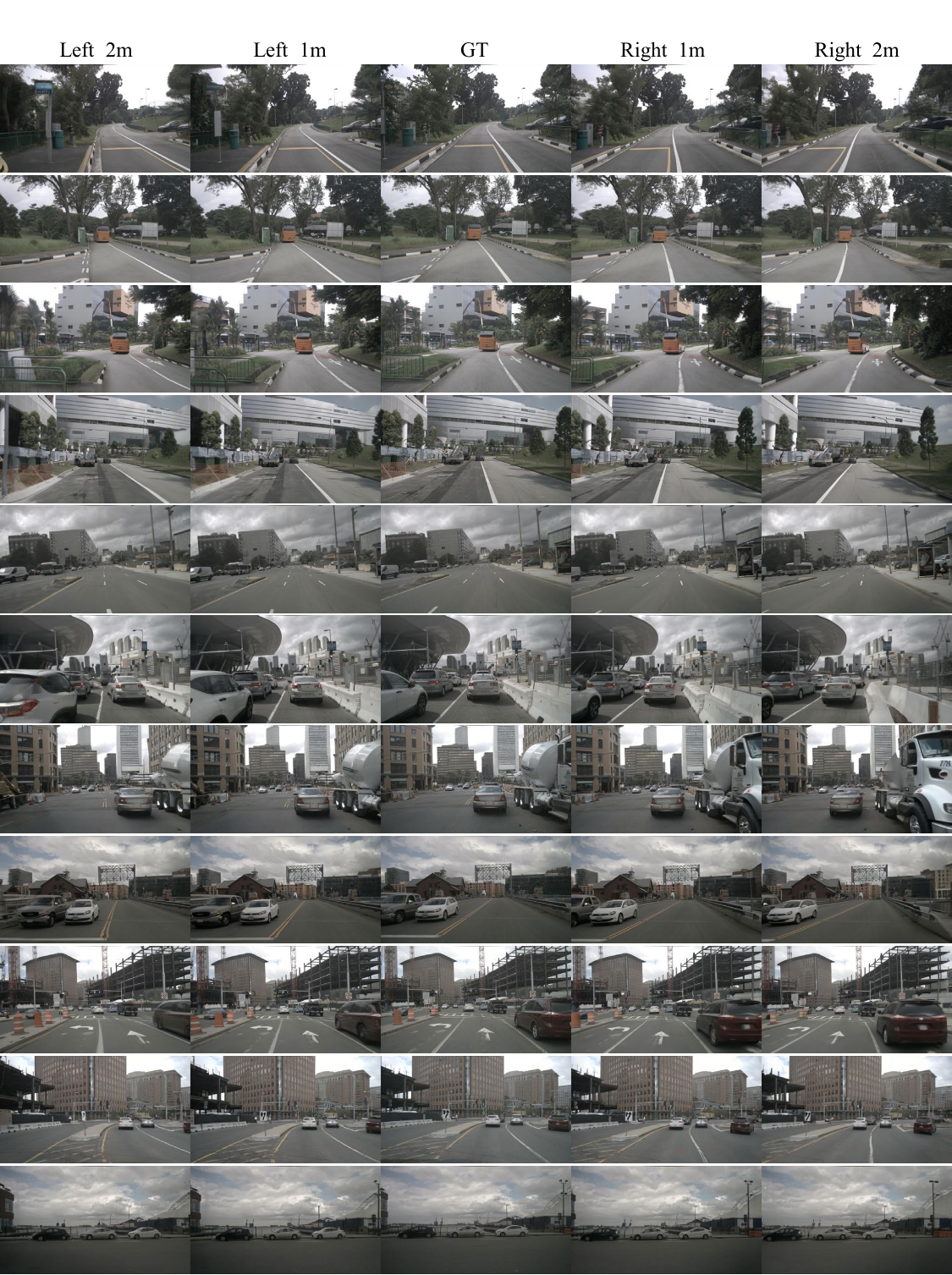}
    \caption{Additional visualizations of spatial novel view synthesis using DiST-S. We present NVS results of DiST-S from various shifted viewpoints, demonstrating our method's ability to generate photorealistic images with high consistency to the original scene.}
    \label{fig_S1}
\end{figure*}

\begin{figure*}[!t]
    \centering
    \includegraphics[width=0.95\linewidth]{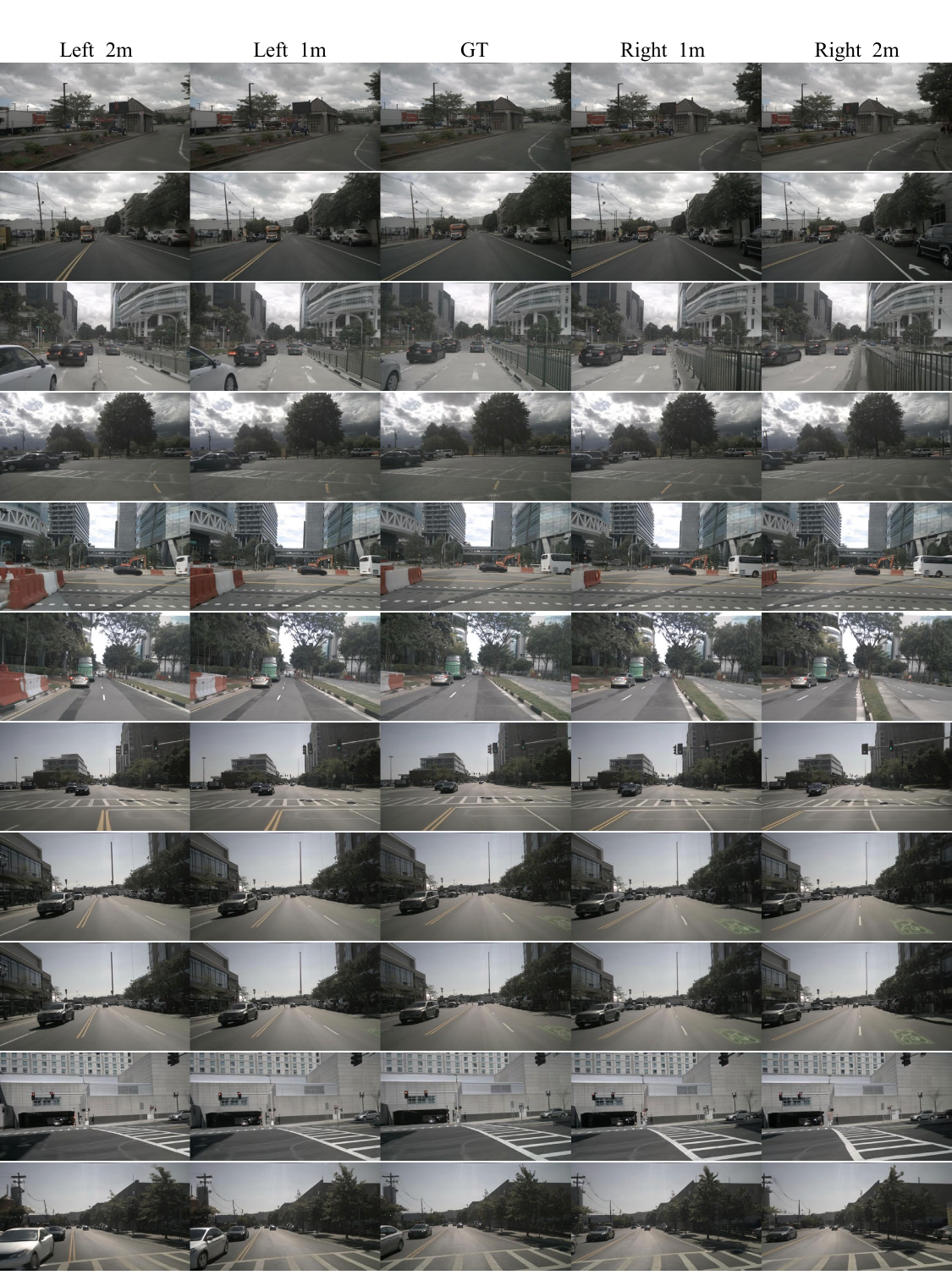}
    \caption{Additional visualizations of spatial novel view synthesis using DiST-S. We present NVS results of DiST-S from various shifted viewpoints, demonstrating our method's ability to generate photorealistic images with high consistency to the original scene.}
    \label{fig_S2}
\end{figure*}

More visualization results about spatial novel view synthesis of DiST-S are provided in Fig.~\ref{fig_S1} and Fig.~\ref{fig_S2}.

\paragraph{Implementation Details of Reconstruction Methods}
To evaluate the effectiveness of our method in novel view synthesis (NVS), we compare it against NeRF-based (EmerNeRF~\cite{yang2023emernerf}) and 3DGS-based (StreetGaussian~\cite{yan2024street}, PVG~\cite{chen2023PVG}, OmniRe~\cite{chen2024omnire}) approaches on 30 scenes from the nuScenes validation dataset. The results are presented in Tab. \ref{tab_NVS} and Fig. \ref{DiST_S_compare_vis}.

For these reconstruction methods, we use all frames in each scene for training. Specifically,
\textbf{EmerNeRF}: We use the official implementation. Since EmerNeRF is configured to train with 100 frames by default on the nuScenes dataset, we split each scene into two subsets, each containing approximately 100 frames.
\textbf{StreetGaussian, PVG, and OmniRe}: We utilize the official code and training configuration provided by OmniRe.

The 30 selected scenes for validation (nuScenes-devkit ~\cite{caesar2020nuscenes} order) are: \textit{11-scene, 12-scene, 13-scene, 14-scene, 36-scene, 75-scene, 79-scene, 83-scene, 84-scene, 87-scene, 88-scene, 90-scene, 91-scene, 92-scene, 214-scene, 257-scene, 259-scene, 261-scene, 262-scene, 410-scene, 412-scene, 414-scene, 436-scene, 439-scene, 442-scene, 443-scene, 444-scene, 445-scene, 446-scene, 447-scene}.

\begin{table}[!t]
\centering
\resizebox{0.99\linewidth}{!}{
\begin{tabular}{l|cccccc}
\toprule
\textbf{Method} &  Abs. Rel. $\downarrow$  & RMSE $\downarrow$ & $\delta < 1.25$ $\uparrow$ & $\delta < 1.25^2$ $\uparrow$  \\
\midrule
\multicolumn{5}{c}{\textit{Multiple Frame LiDAR GT}} \\
\midrule
SD~\cite{wei2023surrounddepth} & 0.27 / 0.28 & 6.50 / 6.59 & 0.67 / 0.63 & 0.87 / 0.85 \\
M$^2$Depth  ~\cite{zou2024m2depth}         &  0.25 / 0.26 & 6.02 / 6.16 & 0.72 / 0.72 & 0.89 / 0.88 \\
DiST-T (Ours-D)               & 0.20 / 0.25 & 5.33 / 5.58 & 0.79 / 0.75 & 0.92 / 0.91 \\
DiST-T (Ours)               & 0.24 / 0.35 & 6.24 / 6.89 & 0.71 / 0.61 & 0.88 / 0.83 \\
Pseudo GT    &  0.13 / 0.20 & 3.46 / 3.31 & 0.84 / 0.76 & 0.95 / 0.93 \\
\bottomrule
\end{tabular}}
\vspace{-5pt}
\caption{Quantitative evaluation of pseudo depth GT and generated depth on nuScense dataset~\cite{caesar2020nuscenes}.
}
\label{tab:depth_generation_results}
\end{table}

\section{Notations}
The notations used in paper are listed in Tab.~\ref{tab:notations}.
\begin{table}[!h]
\centering
\resizebox{0.99\linewidth}{!}{
\begin{tabular}{l l}
\hline
\textbf{Notation} &  \textbf{Description} \\
\hline
$E$ &  VAE encoder \\ 
$D$ &  VAE decoder \\ 
$T$ &  number of frames \\ 
$C$ &  number of cameras \\ 
$\mathbf{I}_{\text{ref},c}$ &  reference image of $c$-th camera \\
$\mathbf{I}_{\text{ref}}$ &  reference images of the all cameras \\
$\mathbf{Z}^{I}$, $\mathbf{Z}^{D}$ &  image, depth latent feature \\ 
$\mathbf{Z}^{I}_{\text{cond}}$, $\mathbf{Z}^{D}_{\text{cond}}$ &  latent feature of image, depth condition \\  
$N$ &  number of blocks in the DiST-T \\ 
$\mathbf{P}_{t}$ &  camera pose at $t$-th frame \\
$\mathbf{B}_{t}$ &  3D bounding boxes at $t$-th frame  \\ 
$\mathbf{A}_{t}$ &  ego trajectory information at $t$-th frame \\ 
$\mathbf{M}_t$ &  map information \\ 
$\mathbf{A}_{\text{ori}}$ &  the original trajectory \\ 
$\mathbf{A}_{\text{novel}}$ &  the novel trajectory \\ 
$V_{t}$, $V_{t+n}$ &  viewpoint of the $t$-th and $t+n$-th frame in $\mathbf{A}_{\text{ori}}$ \\ 
$V_{t^{'}}$ &  viewpoint of the $t$-th frame in $\mathbf{A}_{\text{novel}}$ \\ 
$T^{t \rightarrow t+n}$  & transform matrix from the $V_{t}$ to $V_{t+n}$\\ 
$T^{t \rightarrow t^{'}}$ & transform matrix from the $V_{t}$ to $V_{t^{'}}$ \\
$\tau$ &  laterally shift distance from $\mathbf{A}_{\text{ori}}$ to $\mathbf{A}_{\text{novel}}$\\
$\mathbf{I}_{t+n}^{p}$ &  image condition projected at the $t+n$-th frame \\ 
$\mathbf{D}_{t+n}^{p}$ &  depth condition projected at the $t+n$-th frame \\ 
$\mathbf{C}_{t+n}$ &  \makecell[l]{projected condition at the $t+n$-th frame \\ concatenated from image, depth and valid mask} \\ 
\hline
\end{tabular}}
\caption{Table of notations and descriptions}
\label{tab:notations}
\end{table}

\begin{figure*}[!t]
    \centering
    \includegraphics[width=0.95\linewidth]{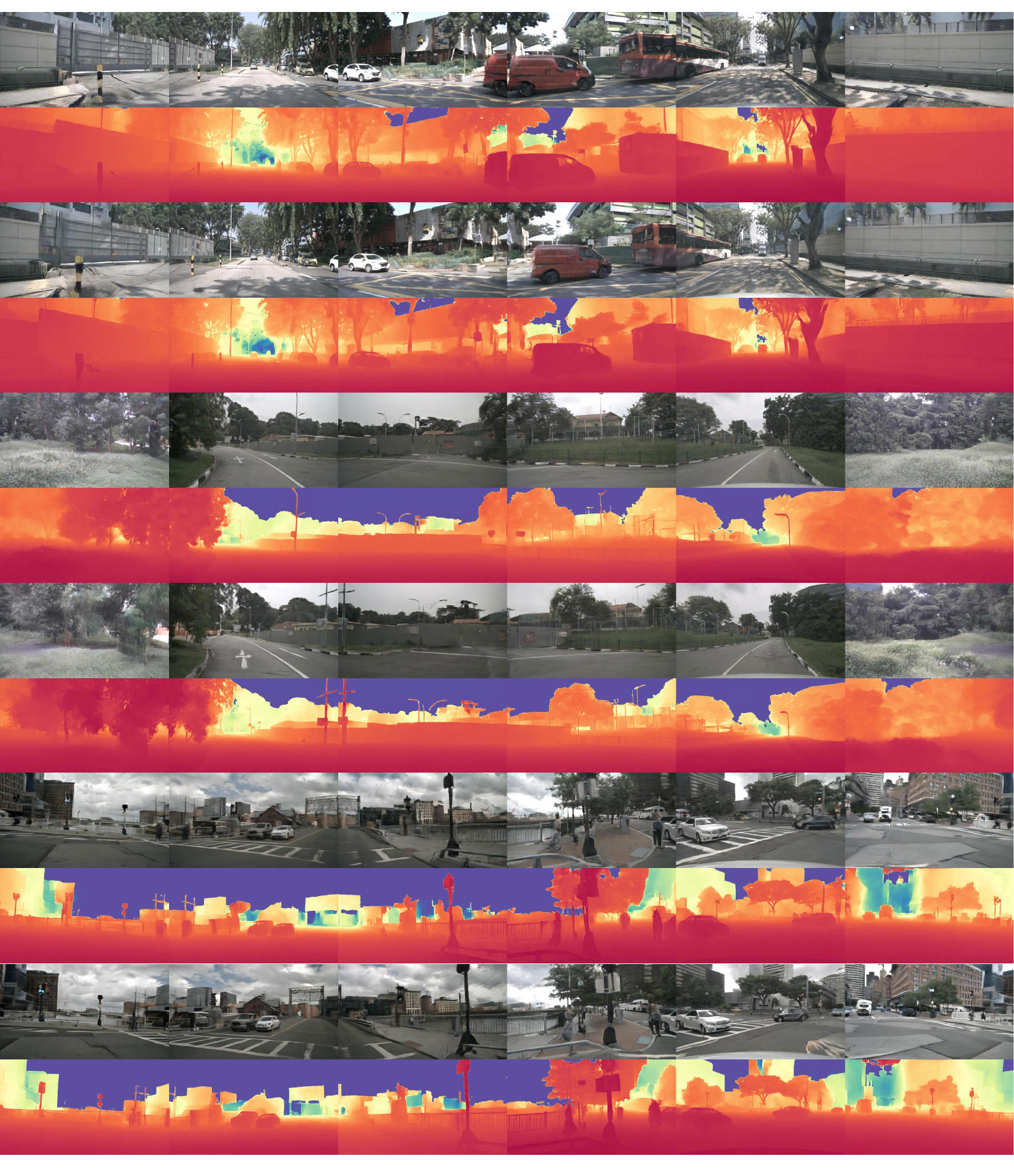}
    \caption{Additional visualizations of video generation using DiST-T. Our model can produce high-quality RGB videos along with corresponding metric depth sequences.}
    \label{fig_T2}
\end{figure*}
\begin{figure*}[!t]
    \centering
    \includegraphics[width=0.95\linewidth]{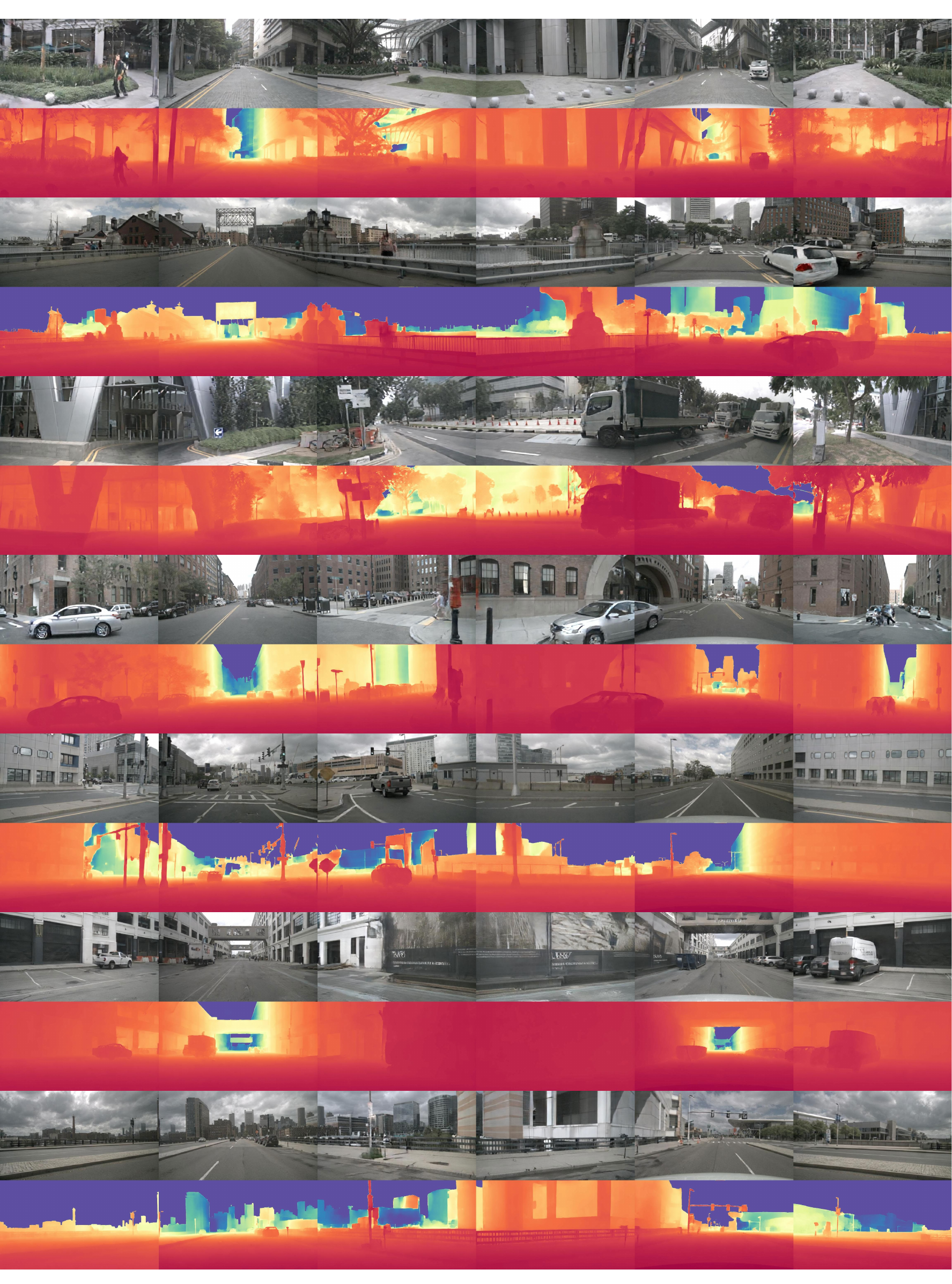}
    \caption{Additional visualizations of video generation using DiST-T. Our model can produce high-quality RGB videos along with corresponding metric depth sequences.}
    \label{fig_T3}
\end{figure*}

\end{document}